%% file: main.tex
\documentclass[twocolumn]{article}

\usepackage{authblk}
\usepackage{amsthm,amsmath}
\usepackage[utf8]{inputenc} 

\usepackage{cite}
\usepackage{amsmath,amssymb,amsfonts}
\usepackage{algorithmic}
\usepackage{graphicx}
\usepackage{textcomp}
\def\BibTeX{{\rm B\kern-.05em{\sc i\kern-.025em b}\kern-.08em
    T\kern-.1667em\lower.7ex\hbox{E}\kern-.125emX}}

\usepackage{hyperref}
\usepackage{tikz}
\usepackage{pgfplots}
\usepgfplotslibrary{fillbetween}
\pgfplotsset{compat=newest}
\usepgfplotslibrary{statistics}
\usetikzlibrary{pgfplots.groupplots}
\usepackage{pgfplotstable}

\usepackage{multirow}
\usepackage{graphicx}
\usepackage{makecell}
\usepackage{booktabs}
\usepackage{array}
\usepackage{comment}

\usepackage{xcolor}
\usepackage[normalem]{ulem}

\pgfplotsset{ every non boxed x axis/.append style={x axis line style=-},
     every non boxed y axis/.append style={y axis line style=-}}

\definecolor{A1}{RGB}{204,102,119}
\definecolor{A2}{RGB}{136,34,85}
\definecolor{A3}{RGB}{136,204,138}
\definecolor{A4}{RGB}{17,119,51}
\definecolor{A5}{RGB}{69,187,247}
\definecolor{A6}{RGB}{51,34,136}
\definecolor{C1}{RGB}{100,100,100}
\definecolor{C2}{RGB}{0,0,0}

\definecolor{black}{RGB}{0,0,0}
\definecolor{intel}{RGB}{0,113,197}

\newcommand{\magenta}[1]{\textcolor{magenta}{#1}}

\newcommand*{\shifttext}[2]{%
  \settowidth{\@tempdima}{#2}%
  \makebox[\@tempdima]{\hspace*{#1}#2}%
}

\pgfplotstableread{results_people_counting_S1_retinaface_gpu.txt}\retinagpuSA
\pgfplotstableread{results_people_counting_S1_mtcnn_gpu.txt}\mtcnngpuSA
\pgfplotstableread{results_people_counting_S1_deepsort_gpu.txt}\deepsortgpuSA
\pgfplotstableread{results_people_counting_S1_trmot_gpu.txt}\trmotgpuSA
\pgfplotstableread{results_people_counting_S1_retinaface_ov.txt}\retinacpuSA
\pgfplotstableread{results_people_counting_S1_mtcnn_ov.txt}\mtcnncpuSA
\pgfplotstableread{results_people_counting_S1_deepsort_ov.txt}\deepsortcpuSA
\pgfplotstableread{results_people_counting_S1_trmot_cpu.txt}\trmotcpuSA
\pgfplotstableread{results_per_sequence_S1_video0.txt}\SAsequenceA
\pgfplotstableread{results_per_sequence_S1_video1.txt}\SAsequenceB
\pgfplotstableread{results_per_sequence_S1_video2.txt}\SAsequenceC
\pgfplotstableread{results_per_sequence_S1_video3.txt}\SAsequenceD
\pgfplotstableread{results_per_sequence_S1_video4.txt}\SAsequenceE
\pgfplotstableread{results_per_sequence_S1_video5.txt}\SAsequenceF
\pgfplotstableread{results_detection_S1_retinaface_gpu.txt}\retinagpuDetSA
\pgfplotstableread{results_detection_S1_mtcnn_gpu.txt}\mtcnngpuDetSA
\pgfplotstableread{results_detection_S1_deepsort_gpu.txt}\deepsortgpuDetSA
\pgfplotstableread{results_detection_S1_trmot_gpu.txt}\trmotgpuDetSA
\pgfplotstableread{results_detection_S1_retinaface_ov.txt}\retinacpuDetSA
\pgfplotstableread{results_detection_S1_mtcnn_ov.txt}\mtcnncpuDetSA
\pgfplotstableread{results_detection_S1_deepsort_ov.txt}\deepsortcpuDetSA
\pgfplotstableread{results_detection_S1_trmot_cpu.txt}\trmotcpuDetSA
\pgfplotstableread{results_people_counting_S2_retinaface_gpu.txt}\retinagpuSB
\pgfplotstableread{results_people_counting_S2_mtcnn_gpu.txt}\mtcnngpuSB
\pgfplotstableread{results_people_counting_S2_deepsort_gpu.txt}\deepsortgpuSB
\pgfplotstableread{results_people_counting_S2_trmot_gpu.txt}\trmotgpuSB
\pgfplotstableread{results_people_counting_S2_retinaface_cpu.txt}\retinacpuSB
\pgfplotstableread{results_people_counting_S2_mtcnn_cpu.txt}\mtcnncpuSB
\pgfplotstableread{results_people_counting_S2_deepsort_cpu.txt}\deepsortcpuSB
\pgfplotstableread{results_people_counting_S2_trmot_cpu.txt}\trmotcpuSB
\pgfplotstableread{results_detection_S2_retinaface_gpu.txt}\retinagpuDetSB
\pgfplotstableread{results_detection_S2_mtcnn_gpu.txt}\mtcnngpuDetSB
\pgfplotstableread{results_detection_S2_deepsort_gpu.txt}\deepsortgpuDetSB
\pgfplotstableread{results_detection_S2_trmot_gpu.txt}\trmotgpuDetSB
\pgfplotstableread{results_detection_S2_retinaface_cpu.txt}\retinacpuDetSB
\pgfplotstableread{results_detection_S2_mtcnn_cpu.txt}\mtcnncpuDetSB
\pgfplotstableread{results_detection_S2_deepsort_cpu.txt}\deepsortcpuDetSB
\pgfplotstableread{results_detection_S2_trmot_cpu.txt}\trmotcpuDetSB
\pgfplotstableread{results_people_counting_S3_retinaface_ov.txt}\retinacpuSC
\pgfplotstableread{results_people_counting_S3_mtcnn_ov.txt}\mtcnncpuSC
\pgfplotstableread{results_people_counting_S3_deepsort_ov.txt}\deepsortcpuSC
\pgfplotstableread{results_people_counting_S3_trmot_cpu.txt}\trmotcpuSC
\pgfplotstableread{results_detection_S3_retinaface_ov.txt}\retinacpuDetSC
\pgfplotstableread{results_detection_S3_mtcnn_ov.txt}\mtcnncpuDetSC
\pgfplotstableread{results_detection_S3_deepsort_ov.txt}\deepsortcpuDetSC
\pgfplotstableread{results_detection_S3_trmot_cpu.txt}\trmotcpuDetSC
\pgfplotstableread{results_people_counting_S4_retinaface_gpu.txt}\retinagpuSD
\pgfplotstableread{results_people_counting_S4_mtcnn_gpu.txt}\mtcnngpuSD
\pgfplotstableread{results_people_counting_S4_deepsort_gpu.txt}\deepsortgpuSD
\pgfplotstableread{results_people_counting_S4_trmot_gpu.txt}\trmotgpuSD
\pgfplotstableread{results_detection_S4_retinaface_gpu.txt}\retinagpuDetSD
\pgfplotstableread{results_detection_S4_mtcnn_gpu.txt}\mtcnngpuDetSD
\pgfplotstableread{results_detection_S4_deepsort_gpu.txt}\deepsortgpuDetSD
\pgfplotstableread{results_detection_S4_trmot_gpu.txt}\trmotgpuDetSD
\pgfplotstableread{results_dropping_retinaface_gpu.txt}\retinaDrop
\pgfplotstableread{results_dropping_mtcnn_gpu.txt}\mtcnnDrop
\pgfplotstableread{results_dropping_deepsort_gpu.txt}\deepsortDrop
\pgfplotstableread{results_dropping_trmot_gpu.txt}\trmotDrop
\pgfplotstableread{results_people_counting_TCOE_S1_retinaface_gpu.txt}\retinagputcoeSA
\pgfplotstableread{results_people_counting_TCOE_S1_mtcnn_gpu.txt}\mtcnngputcoeSA
\pgfplotstableread{results_people_counting_TCOE_S1_deepsort_gpu.txt}\deepsortgputcoeSA
\pgfplotstableread{results_people_counting_TCOE_S1_trmot_gpu.txt}\trmotgputcoeSA
\pgfplotstableread{results_people_counting_TCOE_S1_retinaface_cpu.txt}\retinacputcoeSA
\pgfplotstableread{results_people_counting_TCOE_S1_mtcnn_cpu.txt}\mtcnncputcoeSA
\pgfplotstableread{results_people_counting_TCOE_S1_deepsort_cpu.txt}\deepsortcputcoeSA
\pgfplotstableread{results_people_counting_TCOE_S1_trmot_cpu.txt}\trmotcputcoeSA
\pgfplotstableread{results_people_counting_TCOE_S2_retinaface_gpu.txt}\retinagputcoeSB
\pgfplotstableread{results_people_counting_TCOE_S2_mtcnn_gpu.txt}\mtcnngputcoeSB
\pgfplotstableread{results_people_counting_TCOE_S2_deepsort_gpu.txt}\deepsortgputcoeSB
\pgfplotstableread{results_people_counting_TCOE_S2_trmot_gpu.txt}\trmotgputcoeSB
\pgfplotstableread{results_people_counting_TCOE_S2_retinaface_cpu.txt}\retinacputcoeSB
\pgfplotstableread{results_people_counting_TCOE_S2_mtcnn_cpu.txt}\mtcnncputcoeSB
\pgfplotstableread{results_people_counting_TCOE_S2_deepsort_cpu.txt}\deepsortcputcoeSB
\pgfplotstableread{results_people_counting_TCOE_S2_trmot_cpu.txt}\trmotcputcoeSB
\pgfplotstableread{results_people_counting_TCOE_S3_retinaface_ov.txt}\retinacputcoeSC
\pgfplotstableread{results_people_counting_TCOE_S3_mtcnn_ov.txt}\mtcnncputcoeSC
\pgfplotstableread{results_people_counting_TCOE_S3_deepsort_ov.txt}\deepsortcputcoeSC
\pgfplotstableread{results_people_counting_TCOE_S3_trmot_cpu.txt}\trmotcputcoeSC
\pgfplotstableread{results_people_counting_TCOE_S4_retinaface_gpu.txt}\retinagputcoeSD
\pgfplotstableread{results_people_counting_TCOE_S4_mtcnn_gpu.txt}\mtcnngputcoeSD
\pgfplotstableread{results_people_counting_TCOE_S4_deepsort_gpu.txt}\deepsortgputcoeSD
\pgfplotstableread{results_people_counting_TCOE_S4_trmot_gpu.txt}\trmotgputcoeSD
\pgfplotstableread{results_TCOE_frame_rate_retinaface_gpu.txt}\retinatcoefr
\pgfplotstableread{results_TCOE_frame_rate_mtcnn_gpu.txt}\mtcnntcoefr
\pgfplotstableread{results_TCOE_frame_rate_deepsort_gpu.txt}\deepsorttcoefr
\pgfplotstableread{results_TCOE_frame_rate_trmot_gpu.txt}\trmottcoefr
\pgfplotstableread{results_TCOE_frame_rate_retinaface_ov.txt}\retinatcoefrcpu
\pgfplotstableread{results_TCOE_frame_rate_mtcnn_ov.txt}\mtcnntcoefrcpu
\pgfplotstableread{results_TCOE_frame_rate_deepsort_ov.txt}\deepsorttcoefrcpu
\pgfplotstableread{results_TCOE_frame_rate_trmot_cpu.txt}\trmottcoefrcpu
\pgfplotstableread{results_TCOE_frame_rate_timesegment_peopleOTS_annotation.txt}\OTStimesegment
\pgfplotstableread{results_commercial__retinaface.txt}\commretina
\pgfplotstableread{results_commercial__mtcnn.txt}\commmtcnn
\pgfplotstableread{results_commercial__deepsort.txt}\commdeepsort
\pgfplotstableread{results_commercial__trmot.txt}\commtrmot
\pgfplotstableread{results_commercial__quividi.txt}\commquividi
\pgfplotstableread{results_commercial__meldcx.txt}\commmeldcx
\pgfplotstableread{results_commercial_TCOE_retinaface.txt}\commtcoeretina
\pgfplotstableread{results_commercial_TCOE_mtcnn.txt}\commtcoemtcnn
\pgfplotstableread{results_commercial_TCOE_deepsort.txt}\commtcoedeepsort
\pgfplotstableread{results_commercial_TCOE_trmot.txt}\commtcoetrmot
\pgfplotstableread{results_commercial_TCOE_quividi.txt}\commtcoequividi
\pgfplotstableread{results_commercial_TCOE_meldcx.txt}\commmtcoemeldcx
\pgfplotstableread{results_commercial__retinaface_det.txt}\commretinaDet
\pgfplotstableread{results_commercial__mtcnn_det.txt}\commmtcnnDet
\pgfplotstableread{results_commercial__deepsort_det.txt}\commdeepsortDet
\pgfplotstableread{results_commercial__trmot_det.txt}\commtrmotDet
\pgfplotstableread{results_commercial__quividi_det.txt}\commquividiDet
\pgfplotstableread{results_commercial__meldcx_det.txt}\commmeldcxDet
\pgfplotstableread{results_commercial__facelib_age.txt}\commfacelibAge
\pgfplotstableread{results_commercial__dex_age.txt}\commdexAge
\pgfplotstableread{results_commercial__quividi_age.txt}\commquividiAge
\pgfplotstableread{results_commercial__meldcx_age.txt}\commmeldcxAge
\pgfplotstableread{results_commercial__facelib_gender.txt}\commfacelibGen
\pgfplotstableread{results_commercial__dex_gender.txt}\commdexGen
\pgfplotstableread{results_commercial__quividi_gender.txt}\commquividiGen
\pgfplotstableread{results_commercial__meldcx_gender.txt}\commmeldcxGen
\pgfplotstableread{results_age_S1_facelib_gpu.txt}\facelibgpuAgeSA
\pgfplotstableread{results_age_S1_dex_gpu.txt}\dexgpuAgeSA
\pgfplotstableread{results_age_S1_facelib_ov.txt}\facelibcpuAgeSA
\pgfplotstableread{results_age_S1_dex_ov.txt}\dexcpuAgeSA
\pgfplotstableread{results_age_S2_facelib_gpu.txt}\facelibgpuAgeSB
\pgfplotstableread{results_age_S2_dex_gpu.txt}\dexgpuAgeSB
\pgfplotstableread{results_age_S2_facelib_cpu.txt}\facelibcpuAgeSB
\pgfplotstableread{results_age_S2_dex_cpu.txt}\dexcpuAgeSB
\pgfplotstableread{results_age_S3_facelib_ov.txt}\facelibcpuAgeSC
\pgfplotstableread{results_age_S3_dex_ov.txt}\dexcpuAgeSC
\pgfplotstableread{results_age_S4_facelib_gpu.txt}\facelibgpuAgeSD
\pgfplotstableread{results_gender_S1_facelib_gpu.txt}\facelibgpuGenderSA
\pgfplotstableread{results_gender_S1_dex_gpu.txt}\dexgpuGenderSA
\pgfplotstableread{results_gender_S1_facelib_ov.txt}\facelibcpuGenderSA
\pgfplotstableread{results_gender_S1_dex_ov.txt}\dexcpuGenderSA
\pgfplotstableread{results_gender_S2_facelib_gpu.txt}\facelibgpuGenderSB
\pgfplotstableread{results_gender_S2_dex_gpu.txt}\dexgpuGenderSB
\pgfplotstableread{results_gender_S2_facelib_cpu.txt}\facelibcpuGenderSB
\pgfplotstableread{results_gender_S2_dex_cpu.txt}\dexcpuGenderSB
\pgfplotstableread{results_gender_S3_facelib_ov.txt}\facelibcpuGenderSC
\pgfplotstableread{results_gender_S3_dex_ov.txt}\dexcpuGenderSC
\pgfplotstableread{results_gender_S4_facelib_gpu.txt}\facelibgpuGenderSD

\title{Benchmark for Anonymous Video Analytics}

\author[1]{Ricardo Sanchez-Matilla\thanks{ricardo.sanchezmatilla@qmul.ac.uk}}
\author[1]{Andrea Cavallaro\thanks{a.cavallaro@qmul.ac.uk}}
\affil[1]{Centre for Intelligent Sensing. \\Queen Mary University of London. London. UK}
\date{}

\usepackage{geometry}
\begin{document}

\maketitle

\begin{abstract} 
Out-of-home audience measurement aims to count and characterize the people exposed to advertising content in the physical world. While audience measurement solutions based on computer vision are of increasing interest, no commonly accepted benchmark exists to evaluate and compare their performance. In this paper, we propose the first benchmark for digital out-of-home audience measurement that evaluates the vision-based tasks of audience localization and counting, and audience demographics. The benchmark is composed of a novel, dataset captured at multiple locations and a set of performance measures. Using the benchmark, we present an in-depth comparison of eight open-source algorithms on four hardware platforms with GPU and CPU-optimized inferences and of two commercial off-the-shelf solutions for localization, count, age, and gender estimation. This benchmark and related open-source codes are available at \href{http://ava.eecs.qmul.ac.uk}{http://ava.eecs.qmul.ac.uk}.
\end{abstract}




%



\section{Introduction}

\begin{figure}[!t]
    \centering
    \includegraphics[width=0.9\columnwidth]{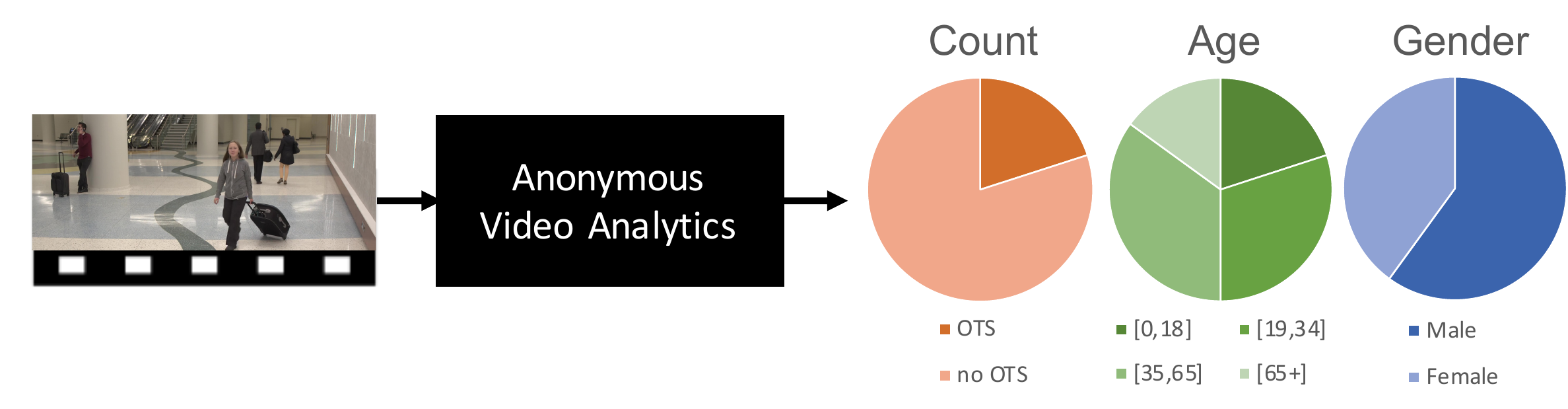}
    \caption{Anonymous Video Analytics for Digital Out-of-home aims to quantify how many people have an Opportunity to See (OTS) a signage and to estimate their demographics.}
    \label{fig:intro}
\end{figure}

Digital out-of-home advertisement is rapidly growing thanks to the availability of affordable, internet-connected smart screens. Anonymous Video Analytics (AVA) aims to enable real-time understanding of audiences exposed to advertisements in order to estimate the reach and effectiveness of each advertisement. AVA {ensures} the preservation of the privacy of audience members by performing inferences and aggregating them directly on edge systems, without recording or streaming raw data (Figure~\ref{fig:intro}). 

AVA relies on person detectors or trackers to localize people and to enable the estimation of audience attributes, such as their demographics. AVA should produce accurate results, be robust to environmental variations and varying illumination. While well-established performance measures exist to evaluate generic computer vision algorithms~\cite{Bernardin2008CLEAR,Idrees2013,Bondi2014,Liu2020}, these measures do not take into account the desirable features of AVA for out-of-home advertisement, such as the opportunity for a person to see the advertisement. In fact, multiple datasets exists to benchmark detection~\cite{Geiger2013,TsungYi2014},  tracking~\cite{Lealtaixe2015,Milan2016,Dendorfer2020,Kristan2018,Kristan2019} and re-identification algorithms~\cite{Gou2017}. Further datasets include KITTI~\cite{Geiger2013}, which focuses on autonomous driving, and Common Objects in Context (COCO)~\cite{TsungYi2014}, which covers object detection, segmentation, and captioning. Moreover, the Robust Vision Challenge~\cite{robustvision} evaluates scene reconstruction, optical flow, semantic and instance segmentation, and depth prediction tasks, the Visual Object Tracking Benchmark~\cite{Kristan2018,Kristan2019} compares single-object tracking algorithms and the
Multiple Object Tracking Benchmark~\cite{Lealtaixe2015,Milan2016,Dendorfer2020} compares multiple-object trackers. {These benchmarks and datasets are designed for scenarios that differ from those for digital out-of-home advertisement and their annotations lack relevant information, such as demographics and attention of the audience, for assessing AVA algorithms.}

Because of the growing importance of digital out-of-home advertisement and the lack of a standard evaluation protocol, in this paper we present the Benchmark for Anonymous Video Analytics\footnote{Benchmark website~\href{http://ava.eecs.qmul.ac.uk}{http://ava.eecs.qmul.ac.uk}.}. This work is the first publicly available benchmark specifically designed to evaluate AVA solutions. The benchmark includes a set of performance measures specifically designed for audience measurement, an online evaluation tool, a novel fully-annotated dataset for digital out-of-home AVA, and open-source baseline algorithms and evaluation codes. The dataset annotations include over a million localization bounding boxes, and age, gender, attention, pose, and occlusion information. We also benchmark eight baseline algorithms: two face detectors, two person trackers, two age estimators, two gender estimators; and two commercial off-the-shelf solutions.

The paper is structured as follows.
Section~\ref{sec:analytics} introduces the main definitions of the work and describes the proposed analytics for AVA;
Section~\ref{sec:performance_measures} presents the performance measures used for benchmarking;
Section~\ref{sec:algorithims} describes the detection, tracking, age, and gender estimation algorithms;
Section~\ref{sec:dataset} introduces the proposed dataset and its annotation;
Section~\ref{sec:benchmark} presents the benchmarking results; and
Section~\ref{sec:conclusion} summarizes the findings of the work.

\subsection{Analytics}
\label{sec:analytics}

\begin{figure}[!t]
    \centering
    \includegraphics[width=0.9\columnwidth]{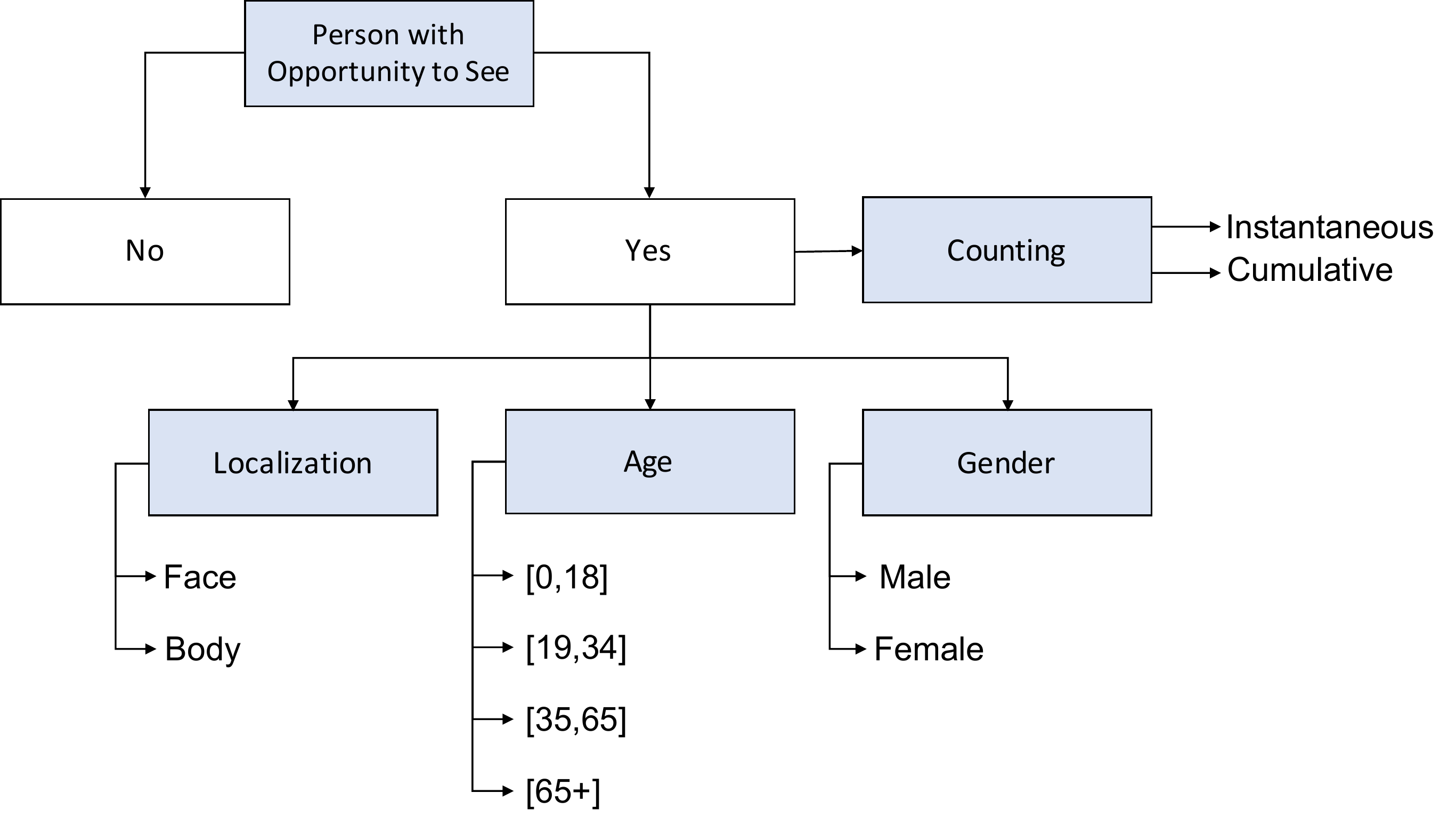}
    \caption{Attributes of people with Opportunity to See a signage that are estimated by Anonymous Video Analytics solutions.}
    \label{fig:taxonomy_people}
\end{figure}

\begin{figure}[!t]
    \centering
    \setlength{\tabcolsep}{1pt}
    \begin{tabular}{ccccc}
        \includegraphics[width=0.19\columnwidth]{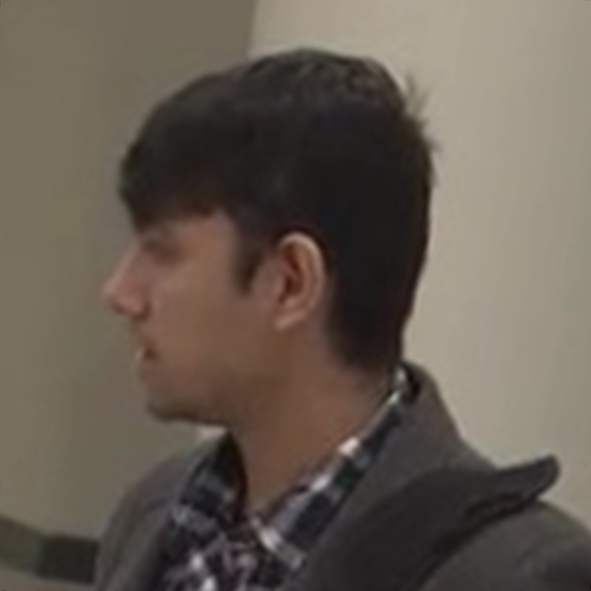} & 
        \includegraphics[width=0.19\columnwidth]{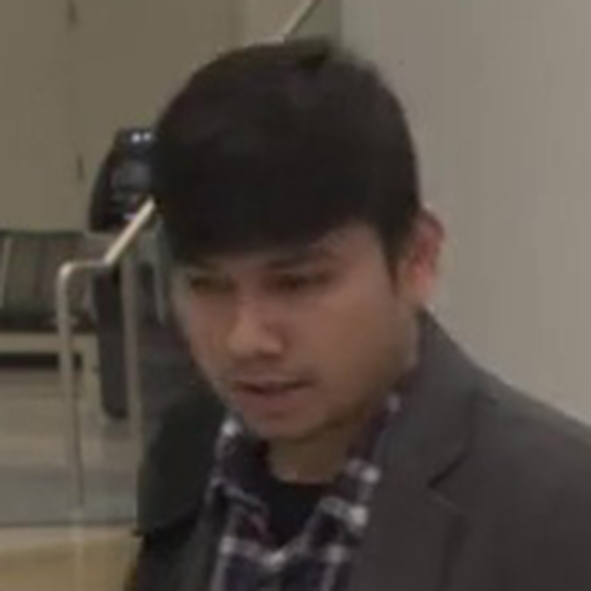} & 
        \includegraphics[width=0.19\columnwidth]{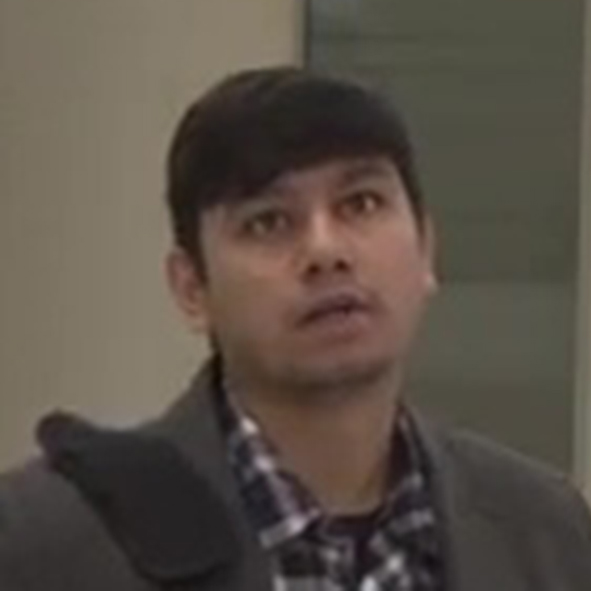} & 
        \includegraphics[width=0.19\columnwidth]{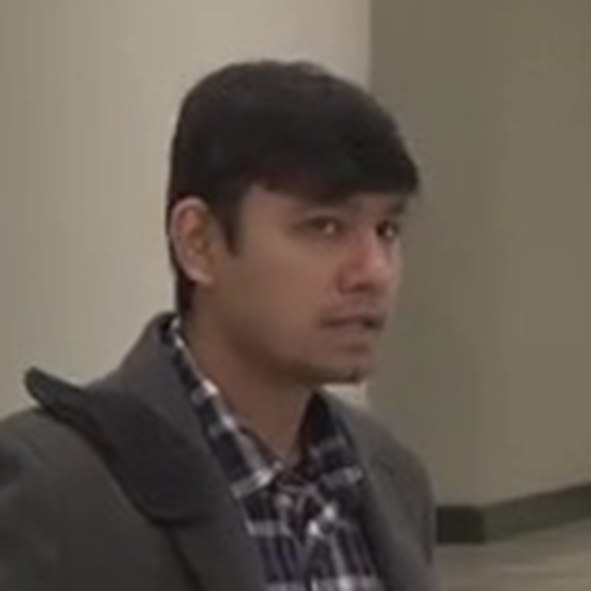} & 
        \includegraphics[width=0.19\columnwidth]{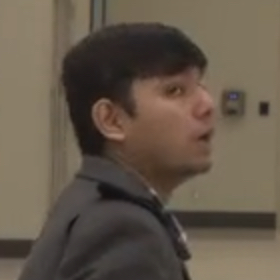}
    \end{tabular}
    \caption{A person is said to have Opportunity to See (OTS) the signage when their face is visible (in any pose from the left to the right profile) and their heading direction is {\em not} opposite to the signage.}
    \label{fig:OTS}
\end{figure}

Let a digital signage be equipped with a system and a camera. The system is a computer that manages the advertisement playback and processes the video of the surroundings of the signage captured by the camera. Let the video $\mathcal{V} = \{ I_{t}\}_{t=1}^{T}$ be composed of $T$ frames $I_{t}$, each with frame index $t$. We consider the following attributes about the people in the video: count, age, and gender (see Figure~\ref{fig:taxonomy_people}). To enable the estimation of the above attributes, we also consider the localization of people in $\mathcal{V}$, namely their position and dimensions in $I_{t}$.
We consider a person in $I_{t}$ to have \textit{Opportunity to See} (OTS) the signage when their face is visible from its left profile to its right profile, and the person is \textit{not} heading opposite to the location of the camera, as shown in Figure~\ref{fig:OTS}. We consider only the attributes of people with OTS.

Let the estimated location and dimensions of person $j \in \mathbb{N}$ with OTS in $I_{t}$ be represented with a bounding box  $\hat{\mathbf{d}}_t^j =[x,y,w,h]$, where $\mathbb{N}$ is the set of the natural numbers. The bounding box is defined by the horizontal, $x$, and vertical, $y$, image coordinates of its top-left corner, and by its width, $w$, and height, $h$. 
The location of person $j$, $\hat{\mathbf{d}}_t^j$, may be represented by their face or their body, and can be estimated with a \textit{detection} or \textit{tracking} algorithm.
While with a detector, the index $j$ may change over time (i.e.~the index $j$ is \textit{not} related to the identity of the person), trackers aim to maintain the index $j$ consistent over time. 

Localization algorithms enable the estimation of the number of people with OTS (counting) at time $t$, $n_t \in \mathbb{N}$, and trackers enable the estimation of the number of \textit{unique}\footnote{Unique means that the algorithm does not repeatedly count the same person over time.} cumulative people with OTS within a time window between $t_1$ and $t_2$, $n_{t_1:t_2} \in \mathbb{N}$.

If $\mathcal{A}$ is a set of age ranges, an age estimation algorithm is expected to determine the age, $a_t^j \in \mathcal{A}$, of a person with OTS, $\hat{\mathbf{d}}_t^j$, with $\mathcal{A}$ defined as:
\begin{equation}
    \mathcal{A} = \{[0,18], [19,34], [35,65], [65+], \text{\textit{unknown}} \}.
\end{equation}
These age ranges have been selected as they are commonly used in audience analytics.

A gender estimation algorithm determines the gender, $g_t^j \in \mathcal{G}$, of each detected person with OTS, $\hat{\mathbf{d}}_t^j$, where
\begin{equation}
    \mathcal{G} = \{\text{\textit{male}}, \text{\textit{female}}, \text{\textit{unknown}} \}
\end{equation} 
is the set of possible classes.

In summary, for each person $j$ with OTS, an AVA solution is expected to produce at each time $t$: $j$, the person index (for trackers, the tracking identity consistent throughout $\mathcal{V}$);
     $\hat{\mathbf{d}}_t^j$, the estimated location of the face and/or body;
    $\hat{a}_t^j \in \mathcal{A}$, the estimated age; and
    $\hat{g}_t^j \in \mathcal{G}$, the estimated gender.



\subsection{Performance measures}
\label{sec:performance_measures}

We introduce a set of performance measures for assessing the accuracy of localization, counting, age and gender estimation. These measures, which are concise and easy to understand by a broad community, enable the evaluation and comparison of AVA algorithms. 


\begin{figure*}[!t]
    \centering
    \includegraphics[width=0.9\textwidth]{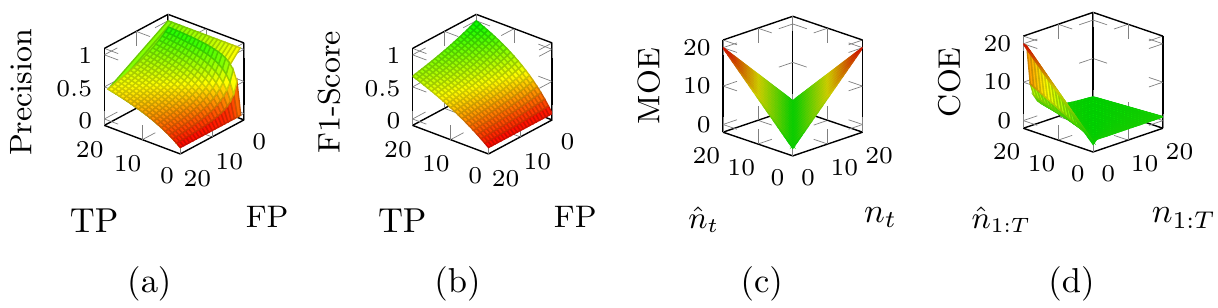}
    \caption{The values of (a) Precision, (b) F1-Score, (c) Mean Opportunity Error (MOE), and (d) Cumulative Opportunity Error (COE) considering a case with 20 actual positives. Precision and F1-Score are a function of true positives (TP) and false positives (FP).
    The MOE is shown as a function of the estimated ($\hat{n}_t$) and actual ($n_t$) number of people with OTS. The COE is shown as a function of the estimated ($\hat{n}_{1:T}$) and actual ($n_{1:T}$) cumulative number of people with OTS for the whole video.
    }
    \label{fig:sample_chars}
\end{figure*}

\textbf{Localization}

We evaluate the localization performance based on \textit{precision} ($\text{P}$), \textit{recall} ($\text{R}$), and \textit{F1-Score} ($\text{F}$)~\cite{Liu2020}, which are defined based on true positives {(TP)}, false positives {(FP)}, and false negatives {(FN)}.
{\textit{Precision} ($\text{P}$) is the ratio between correct estimations and the total number of estimations.
\textit{Recall} ($\text{R}$) is the ratio between correct estimations and the total number of actual occurrences.
\textit{F1-Score} ($\text{F}$) is the harmonic mean of $\text{P}$ and $\text{R}$.}

\begin{figure}[!t]
    \centering
    \includegraphics[trim=0 300 0 0, clip, width=0.9\columnwidth]{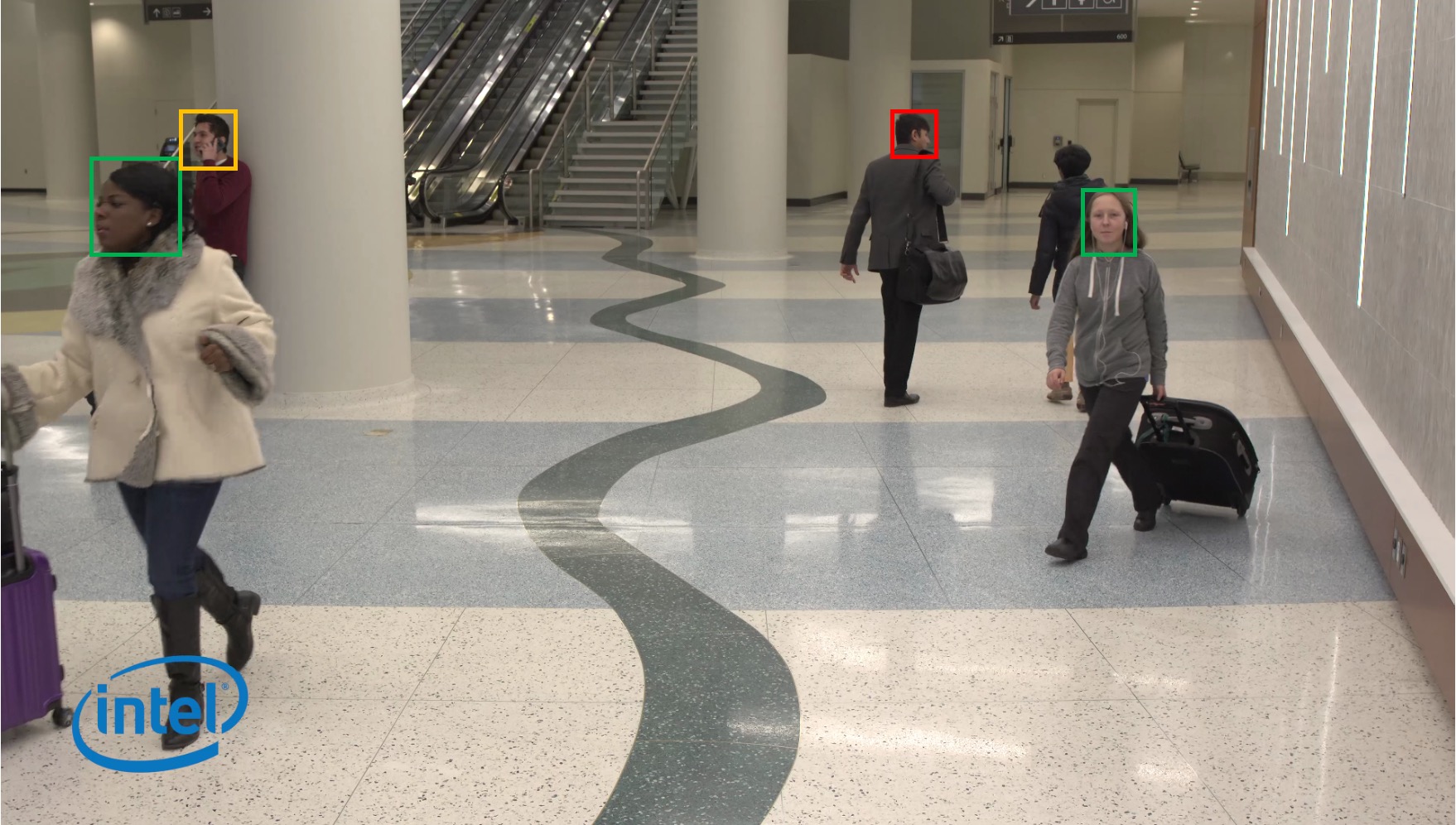}
    \caption{Sample result and evaluation for people counting. Green, yellow, and red bounding boxes indicate true positives, false negatives, and false positive, respectively.
    Assuming that the frame is the first in a video, $t=1$, the actual number of people is five ($p_1=5$), from which only three have OTS ($n_1=3$). The measures are $\text{MOE}=0$, $\text{MPE}=2$; and $\text{COE}=0$. Note that the person marked with a red bounding, even that his face is visible, we consider he has no OTS as he is walking in a direction opposite to the location of the camera. The frame is cropped. 
    }
    \label{fig:visual_sample}
\end{figure}

For localization, we define TP, FP, and FN {based on} the intersection over union (IOU) operator between estimations, $\hat{\mathbf{d}}^j_t$, and annotations, $\mathbf{d}^j_t$: 
\begin{equation}
    \text{IOU}(\hat{\mathbf{d}}^j_t,\mathbf{d}^j_t) = 
    \frac{\hat{\mathbf{d}}^j_t \cap \mathbf{d}^j_t}{\hat{\mathbf{d}}^j_t \cup \mathbf{d}^j_t},
\end{equation}
where $\cap$ and $\cup$ are the intersection and the union operators, and $\text{IOU} \in [0,1]$. 
An example for face localization{, with highlighted TP, FP and FN detections} is given in Figure~\ref{fig:visual_sample}. 

We consider the variation of R for different person-signage distances.
Let $A(\cdot)$ be a function that computes the area in pixels for bounding box $\mathbf{d}_t^j$, and $p_{\text{dist}}^{a}$ be the $a$-th percentile of all the bounding boxes in the video.
We define two bands for the person-signage distance, namely \textit{close} when $A(\mathbf{d}_t^j) \ge p_{\text{dist}}^{50}$ and \textit{far} when $A(\mathbf{d}_t^j) < p_{\text{dist}}^{50}$. We assume that people closer to the camera are annotated by a larger bounding box than for those who are farther.
%

We also consider the variation of R in presence of occlusions. 
Let $o_t^j \in \{\text{\textit{non occluded}}, \text{\textit{partially occluded}}, \allowbreak \text{\textit{heavily occluded}}\}$ be the annotated occlusion. The three occlusion bands are \textit{non occluded} when the annotation is not occluded; \textit{partially occluded} when the annotated area is occluded less than 50\%; and \textit{heavily occluded} when the annotated area is occluded more or equal than 50\%.
%

{Note that we only report R, and not P, for person-signage distance and occlusion as false positives (necessary to compute P) cannot be unequivocally estimated when the annotations are divided in regions such as \textit{far}/\textit{close} (for distance) or \textit{non occluded}/\textit{partially occluded}/\textit{heavily occluded} (for occlusion). For instance, a \textit{far}/\textit{close} estimation might not match with any \textit{far}/\textit{close} annotation but this does not necessarily imply to be a false positive as it might be matching with a \textit{close}/\textit{far} annotation.}

\begin{figure*}[!t]
    \centering 
    \includegraphics[width=1.0\textwidth]{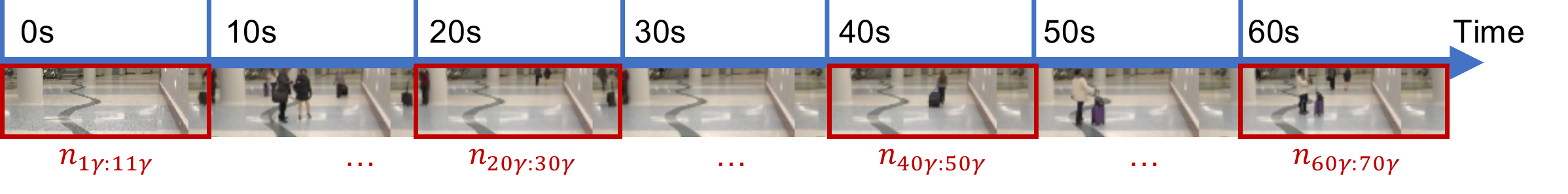}
    \caption{Sample Temporal Cumulative Opportunity Error (TCOE) computation for a video of 70 seconds, with an example of 10-second segment, $D=10 \gamma$, where $\gamma$ is the frame rate of the video in frames per second. When computing TCOE, all segments of $D=\{10,20,30,60,90,120\} \gamma$ frames (i.e.~segments of 10, 20, 30, 60, 90, 120~second duration) within the video are used.}
    \label{fig:TCOE}
\end{figure*}

\textbf{Counting}

We quantify the performance of the localization algorithms for the task of people counting with the following performance measures: Mean Opportunity Error (MOE), Cumulative Opportunity Error (COE), and Temporal Cumulative Opportunity Error (TCOE).

The Mean Opportunity Error (MOE) quantifies the ability of an algorithm to count people with OTS at a specific time $t$, $\hat{n}_t$, and it is calculated with respect to the actual number of people with OTS at $t$, $n_t$:
\begin{equation}
    \text{MOE} =  \frac{1}{T} \sum_{t=1}^{T}{{|\hat{n}_t-n_t|}}.
\end{equation}
$\text{MOE} \ge 0$ (Figure~\ref{fig:sample_chars}c) and its optimal value is $\text{MOE}=0$. We analyze how MOE varies with the person-signage distance, as in localization, and when the input video frame rate is reduced. We show a visual sample of MOE in Figure~\ref{fig:visual_sample}.

The Cumulative Opportunity Error (COE) quantifies the ability of an algorithm to count unique people with OTS and it is calculated with respect to the actual cumulative number of people with OTS for the whole video, $n_{1:T}$:
\begin{equation}
    \text{COE} =  \frac{|\hat{n}_{1:T}-n_{1:T}|}{\text{max}(n_{1:T},1)},
\end{equation}
where $\text{max}(\cdot)$ is the max operation. $\text{COE} \ge 0$ and its optimal value is $\text{COE}=0$. This performance measure is normalized with respect to the actual cumulative number of people; therefore, COE indicates the ratio of error with respect to the actual cumulative number of people.

The Temporal COE (TCOE) quantifies the ability of an algorithm to count unique people with OTS over temporal segments of generic duration (e.g.~10~second duration), and it is calculated with respect to the cumulative number of unique people with OTS:
%
%
\begin{equation}
    \text{TCOE}_{D,T} =  \frac{1}{|\mathcal{T}_{D,T}|} \sum_{\forall t \in \mathcal{T}_{D,T}}{ { |\hat{n}_{t:t+D}-n_{t:t+D}|} },
\end{equation}
where $\mathcal{T}_{D,T}=\{1,2,3,\dots,T-D\}$ is the set of the initial frame of each segment, $D < T$ is the duration in frames of the segments, and $|\mathcal{T}_{D,T}|$ is the total number of segments. We consider values that correspond to the typical duration of digital out-of-home advertisements $D=\{10,20,30,60,90,120\} \, \gamma$, where $\gamma$ is the video frame rate in~fps (i.e.~segments of 10, 20, 30, 60, 90, 120~second duration).
$\text{TCOE}$ considers all possible $D$-frame segments within the video. We show an example for 10~second segments in Figure~\ref{fig:TCOE}.
$\text{TCOE}_{D,\mathcal{T}} \ge 0$ and its optimal value is $\text{TCOE}_{D,\mathcal{T}}=0$. Note that when $D=T$, TCOE equals COE.

To quantify whether algorithms estimate all people in the field of view but fail to only estimate the ones with OTS, we define two accessory measures, Mean People Error (MPE) and Cumulative Person Error (CPE). If $p_t$ is the number of people at $t$ and $p_{1:T}$ is the number of people in the whole video, then
\begin{equation}
    \text{MPE} = \frac{1}{T} \sum_{t=1}^{T}{|\hat{n}_t-p_t|},
\end{equation}
%
and
%
\begin{equation}
    \label{eq:cpe}
    \text{CPE} = \frac{|\hat{n}_{1:T}-p_{1:T}|}{\text{max}(p_{1:T},1)}.
\end{equation}
%

\begin{table}[!t]
    \centering
    \caption{Example of classification results for a given attribute (i.e.~gender) using three different algorithms. Note that an algorithm that can output \textit{unknown} produces higher performance measures than one that commits the same amount of errors (e.g. algorithm B vs algorithm A).  For gender estimation and for the class \textit{female},  a true positive is a female estimated as female; a false negative is a female estimated as male; and a false positive is a male estimated as female.}
    \label{tab:classification}
    \resizebox{\columnwidth}{!}{%
    \begin{tabular}{lrrr}
      \specialrule{1.2pt}{0.2pt}{1pt}
         & \multicolumn{1}{c}{Algorithm A} & \multicolumn{1}{c}{Algorithm B} & \multicolumn{1}{c}{Algorithm C} \\
        \specialrule{1.2pt}{0.2pt}{1pt}
        True positives & 100 & 100 & 110 \\
        False negatives & 51 & 51 & 51\\
        False positives & 24 & 14 & 14\\
        Unknown & 0 & 10 & 0 \\
        \hline
        Precision & 0.81 & 0.88 & 0.89 \\
        Recall & 0.66 & 0.66 & 0.68 \\
        F1-Score & 0.73 & 0.76 & 0.77 \\
      \specialrule{1.2pt}{0.2pt}{1pt}
    \end{tabular}
    }
\end{table}

\begin{figure}
    \centering
    \includegraphics[width=0.9\columnwidth]{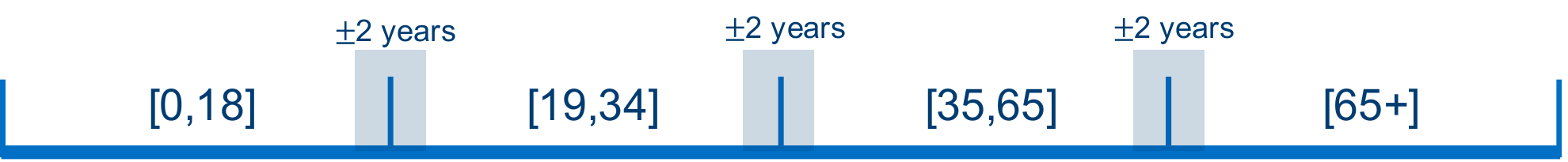}
    \caption{Age ranges and their overlap of $\pm$ 2 years for error calculation.}
    \label{fig:age}
\end{figure}

\textbf{Attributes}

We compute the per-class precision, recall, and F1-Score for age and gender estimation. We consider the variation of F for different person-signage distances and for different occlusion levels. 


We give now a few examples of the definitions of TP, FP, and FN for each attribute. For age estimation and the class [19,34], a TP is a correct age estimation. To relax the hardness of the age ranges, we consider overlapping age ranges with $\pm$ 2 years, as shown in Figure~\ref{fig:age}, (e.g. an estimation of a person of 17 years will be a true positive if the actual age of the person is in [0,18], or [19,34]); a FP is an incorrect age estimation for a person from another age-class; and a FN is an incorrect estimation of the age of a person that belongs to the class.
For gender estimation and the class \textit{female}, a TP is a female estimated as female; a FP is a male estimated as female; and a FN is a female estimated as male.

For algorithms able to output \textit{unknown} as a possible class, the corresponding estimations will contribute neither as TP, FP nor FN.
We show an example of attribute evaluation in Table~\ref{tab:classification}.

\section{Methods}
\label{sec:algorithims}

\begin{table*}[!t]
    \centering
    \setlength{\tabcolsep}{2pt}
    \caption{Anonymous Video Analytics algorithms for localization, age, and gender estimation. Algorithms \textit{count performance} and \textit{speed} are assessed, on a scale from 1 to 3 (the higher, the better) and regardless of the system used, based on generic algorithmic concepts (e.g.~tracking outperforms detection for cumulative counts) and not on experimental results.  KEY - OTS: Opportunity to See.}
    \label{tab:algorithms_summary}
    \resizebox{0.95\textwidth}{!}{%
    \begin{tabular}{llllcccccccccc}
    \specialrule{1.2pt}{0.2pt}{1pt}
    \textbf{Alias} & \multicolumn{1}{c}{\textbf{Ref.}} & \multicolumn{1}{c}{\textbf{Name}} & \multicolumn{1}{c}{\textbf{Person part}} & \multicolumn{2}{c}{\textbf{Localization}} & \multicolumn{1}{c}{\textbf{Age}} & \multicolumn{1}{c}{\textbf{Gender}} & \multicolumn{2}{c}{\textbf{Implementation}} & \multicolumn{3}{c}{\textbf{Count performance}} & \textbf{Speed} \\
    \cmidrule(lr){5-6}\cmidrule(lr){9-10}\cmidrule(lr){11-13}
     &  &  &  & \multicolumn{1}{c}{Detection} & \multicolumn{1}{c}{Tracking} &  &  & \multicolumn{1}{c}{GPU} & \multicolumn{1}{c}{CPU} & \multicolumn{2}{c}{Instantaneous} & Cumulative &  \\
     \cmidrule(lr){11-12}
     &  &  &  &  &  &  &  &  &  & People & OTS &  &  \\
    \specialrule{1.2pt}{0.2pt}{1pt}
    A1 & \cite{Deng2020} & RetinaFace & Face & \checkmark &  &  &  & \checkmark & \checkmark & + & +++ & + & +++ \\
    A2 & \cite{Zhang2016} & MTCNN & Face & \checkmark &  &  &  & \checkmark & \checkmark & + & +++ & + & +++ \\
    A3 & \cite{Wojke2017} & DeepSORT & Person &  & \checkmark &  &  & \checkmark & \checkmark & +++ & + & +++ & ++ \\
    A4 & \cite{Wang2019} & TRMOT & Person &  & \checkmark &  &  & \checkmark &  & +++ & + & +++ & ++ \\
    A5 & \cite{Ayoubi2020} & FaceLib & Face & - & - & \checkmark & \checkmark & \checkmark & \checkmark & - & - & - & +++ \\
    A6 & \cite{Rothe2015} & DEX & Face & - & - & \checkmark & \checkmark & \checkmark & \checkmark & - & - & - & +++ \\ \hline
    C1 & \multicolumn{2}{l}{Commercial-1} & Face &  & \checkmark & \checkmark & \checkmark &  & \checkmark & + & +++ & +++ & ++ \\
    C2 & \multicolumn{2}{l}{Commercial-2} & Face &  & \checkmark & \checkmark & \checkmark &  & \checkmark & +++ & +++ & +++ & ++ \\
    \specialrule{1.2pt}{0.2pt}{1pt}
    \multicolumn{14}{l}{A4 was not compatible with OpenVINO optimization at submission time, thus we run A4 on CPU without optimization.}
    \end{tabular}
    }
\end{table*}


{Algorithms must be accurate and perform in real-time for being suitable to generate reliable and useful AVA.
Therefore, we select algorithms for benchmarking that obtain close to state-of-the-art results, that are causal (i.e. they only need past and present information), and that are able to perform close to real-time in the defined settings. We select algorithms that are compatible with GPU and OpenVINO\footnote{OpenVINO framework is available at \href{https://docs.openvinotoolkit.org/}{https://docs.openvinotoolkit.org/}.} optimization for fast CPU computation.}

We use \textit{RetinaFace}~\cite{Deng2020} and \textit{Joint Face Detection and Alignment using Multi-task Cascaded Convolutional Networks} (MTCNN)~\cite{Zhang2016} as detection algorithms; 
\textit{Simple Online Real-time Tracking with a Deep Association Metric} (DeepSORT)~\cite{Wojke2017}, and \textit{Towards Real-Time Multi-Object Tracking} (TRMOT)~\cite{Wang2019} as trackers; and
\textit{FaceLib}~\cite{Ayoubi2020} and \textit{Deep EXpectation of apparent age from a single image} (DEX)~\cite{Rothe2015} as age and gender estimators.
In addition to the above baseline algorithms, we benchmark two commercial solutions, \textit{Commercial-1} (C1) and \textit{Commercial-2} (C2), which we maintain anonymous.
Localization, age, and gender estimation algorithms for AVA are described next and summarized in Table~\ref{tab:algorithms_summary}.

Algorithm~1 (A1), RetinaFace~\cite{Deng2020}, is a face detector with a single-stage pixel-wise dense localization at multiple scales that uses joint extra-supervised and self-supervised multi-task learning. The algorithm predicts a face score, face box, five facial landmarks, and their relative 3D position, using input images resized to a resolution of 640$\times$640 pixels. The algorithm is trained on the WIDER FACE dataset~\cite{Yang2016}.

Algorithm~2 (A2), MTCNN~\cite{Zhang2016}, is a face detector with a cascaded structure and three stages of deep convolutional networks that use the correlation between the face bounding box and the landmark localization to perform both tasks in a coarse-to-fine manner. 
The first stage is a shallow network that generates candidate windows.
The second stage rejects false positive candidate windows.
The third stage is a deeper network that outputs the locations and facial landmarks.
The learning process uses an online hard sample mining strategy that unsupervisedly improves the performance. 
The algorithm is trained on the WIDER FACE dataset~\cite{Yang2016}.

Algorithm~3 (A3), DeepSORT~\cite{Wojke2017}, is a  multi-object tracker that combines the detector YOLOv3~\cite{Redmon2018} and the tracker SORT~\cite{Bewley2016}. 
YOLOv3 detects the body of people with a single neural network that divides the image into regions and predicts bounding boxes and probabilities for each region. These bounding boxes are weighted by the predicted probabilities. The predictions are informed by the global context of the image.
With respect to the tracking module, DeepSORT uses Kalman Filter and the Hungarian algorithm for performing association of detections over time. In addition, DeepSORT employs a convolutional neural network, trained to discriminate people, that combines appearance and motion information.
 
Algorithm~4 (A4), TRMOT~\cite{Wang2019}, is a multi-object tracker based on the Joint Detection and Embedding (JDE) framework. JDE is a single-shot shared deep neural network that simultaneously learns detection and appearance features of the predictions. The algorithm is based on Feature Pyramid Network~\cite{Lin2017} that makes predictions at multiple scales. Then, embedding features are up-sampled and fused with the feature map from higher feature maps levels by using skip connections to improve the tracking accuracy for people far from the camera (i.e.~small bounding boxes). The JDE framework learns to generate predictions and features simultaneously.

Algorithm~5 (A5), FaceLib~\cite{Ayoubi2020}, is an open-source repository for face detection, facial expression recognition, and age and gender estimation. The age and gender estimation modules use as input the true positive detections generated by RetinaFace (A1), and they use a ShuffleNet V2 with 1.0x output channels~\cite{Ningning2018} as architecture. FaceLib is trained on the UTKFace dataset~\cite{Zhifei2017}.

Algorithm~6 (A6), DEX~\cite{Rothe2015}, estimates the apparent age and gender using as input the true positive detections generated by RetinaFace (A1). DEX uses an ensemble of 20 networks on the detected face and it does not require the use of facial landmarks. This work introduces and uses the IMDB-WIKI, a large public dataset of face images with age and gender annotations.

Commercial off-the-shelf~1 (C1) is composed of a frontal face detector based on a cascade detector that uses a variety of gray-scale local features, a tracker designed for frontal, eye-level images; and a gender and age classifier based on regression trees.

Commercial off-the-shelf~2 (C2) uses a tracker based on a Kalman Filter for tracking people specifically designed to be robust to occlusions; and a gender and age classifier based on deep learning.

\begin{figure*}[!t]
    \centering
    \setlength{\tabcolsep}{1pt}
    \begin{tabular}{ccccccc}
        \includegraphics[width=0.13\textwidth]{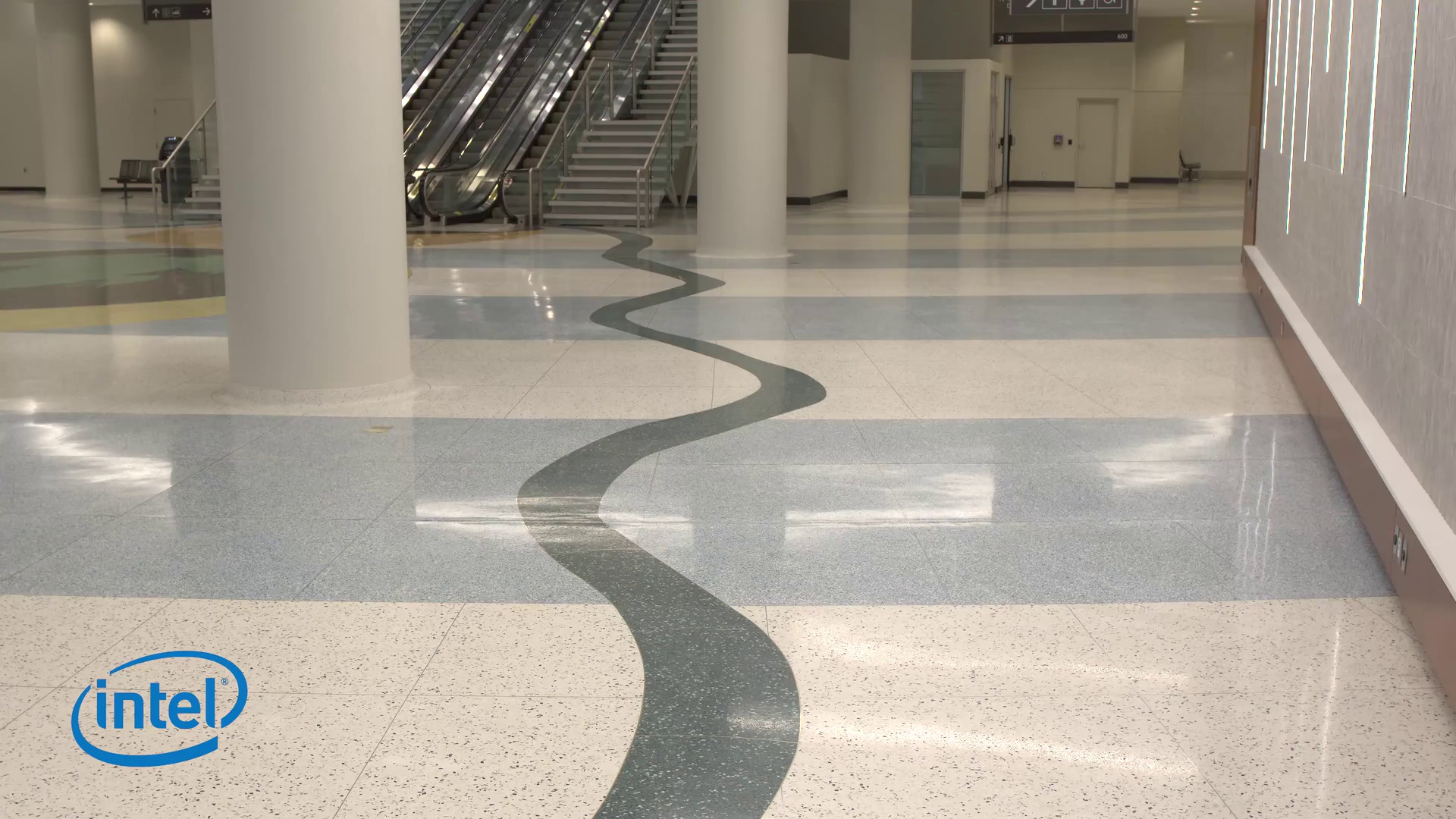} &
        \includegraphics[width=0.13\textwidth]{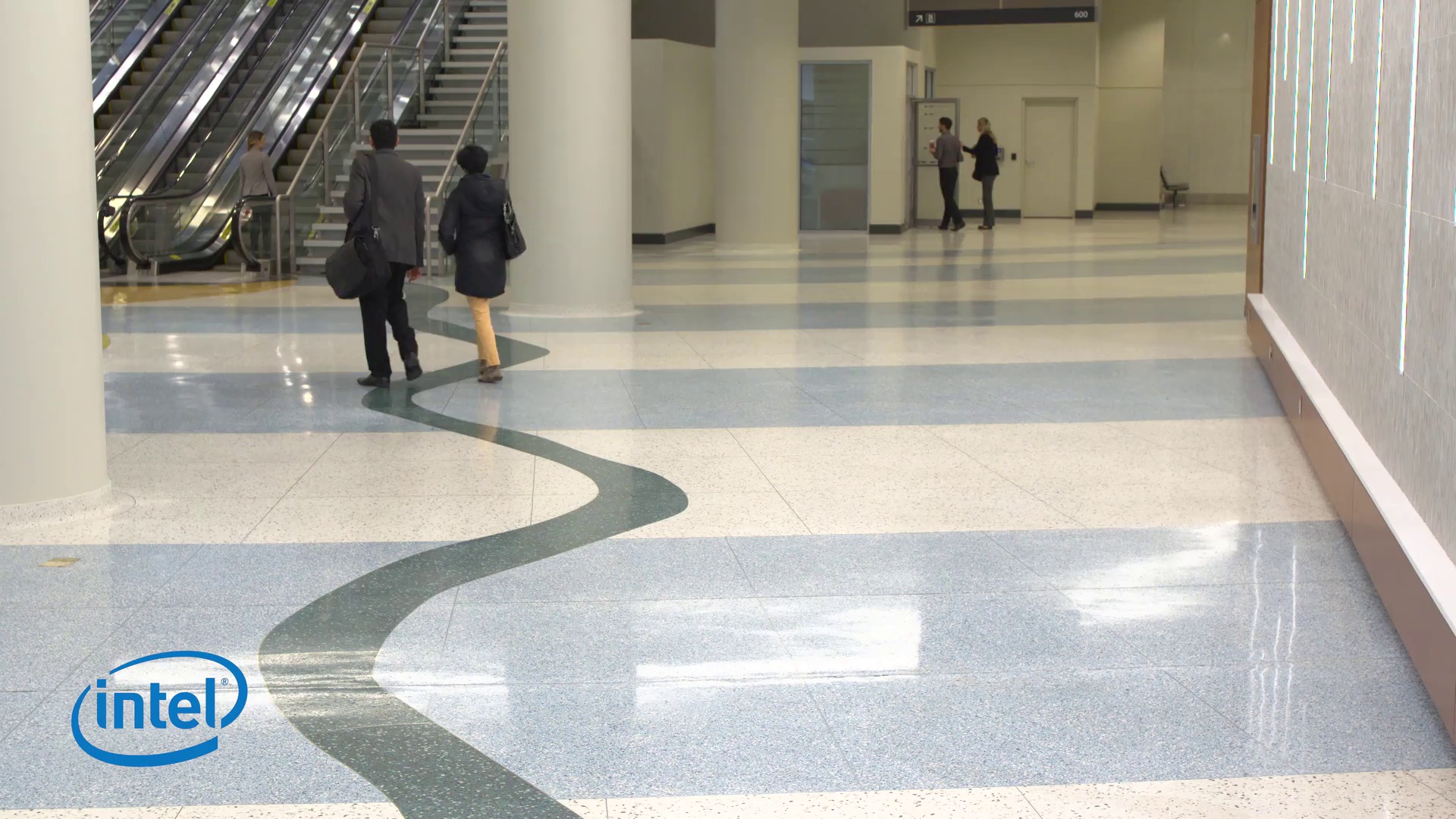} &
        \includegraphics[width=0.13\textwidth]{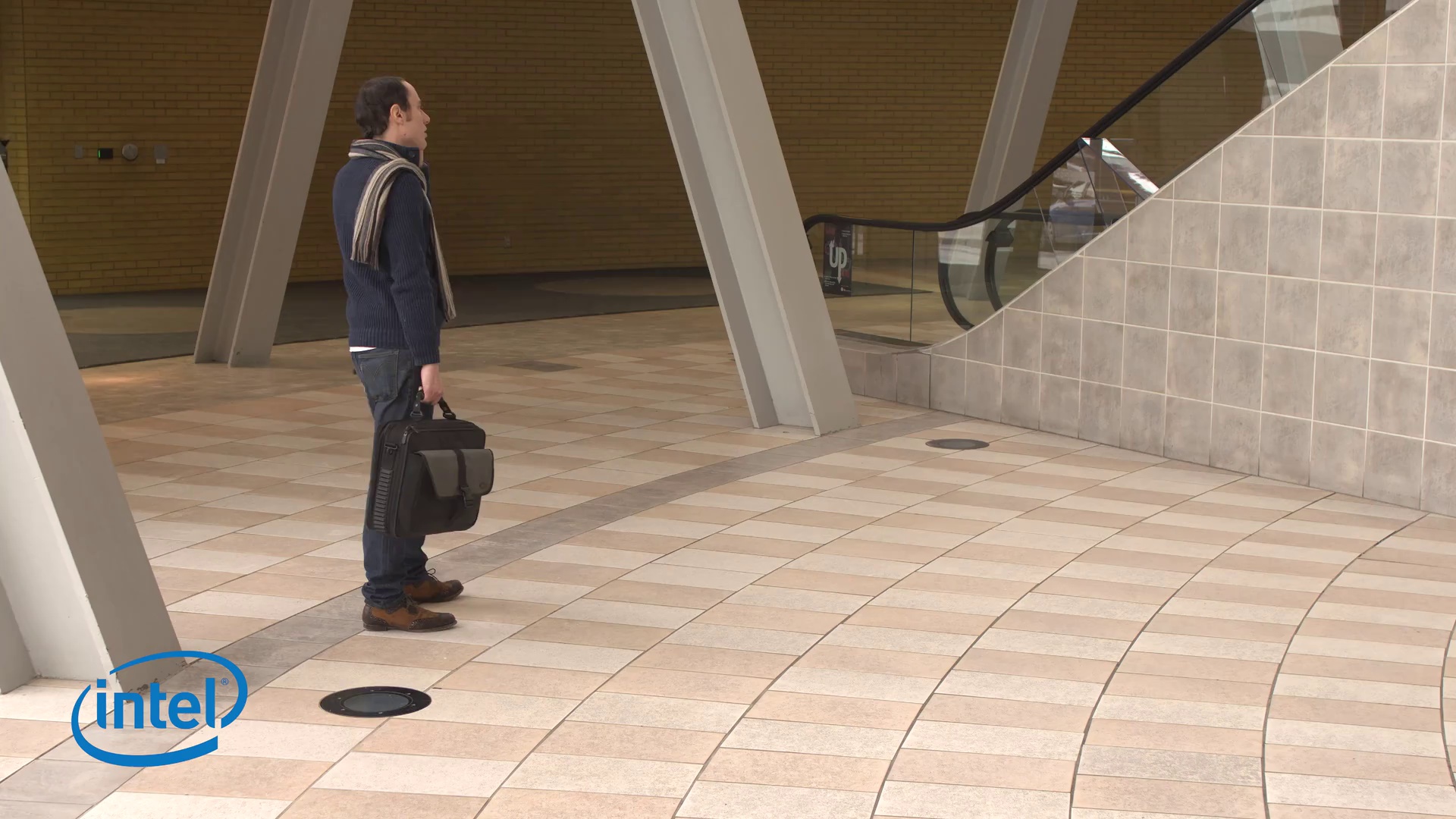} &
        \includegraphics[width=0.13\textwidth]{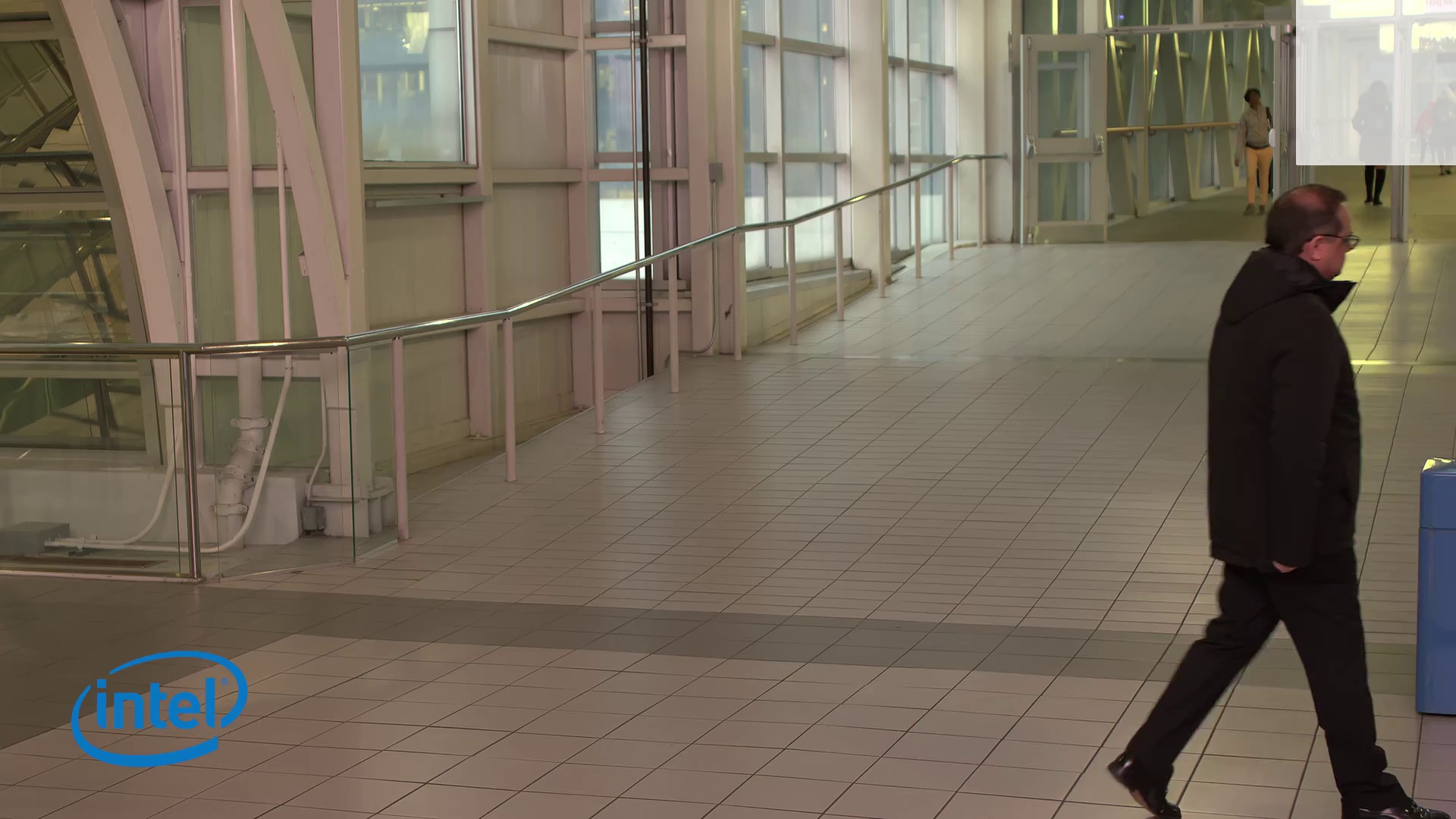} &
        \includegraphics[width=0.13\textwidth]{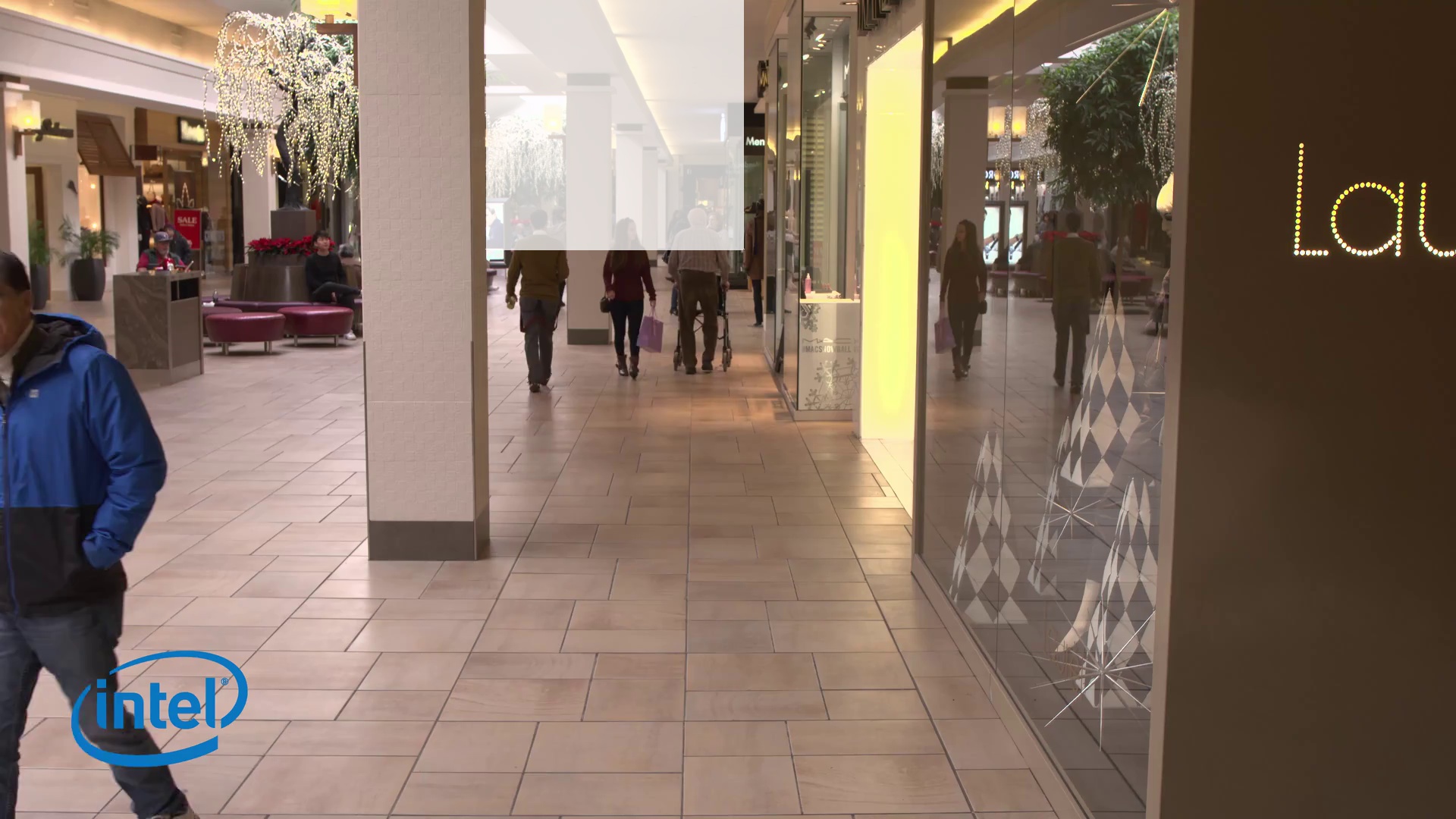} &
        \includegraphics[width=0.13\textwidth]{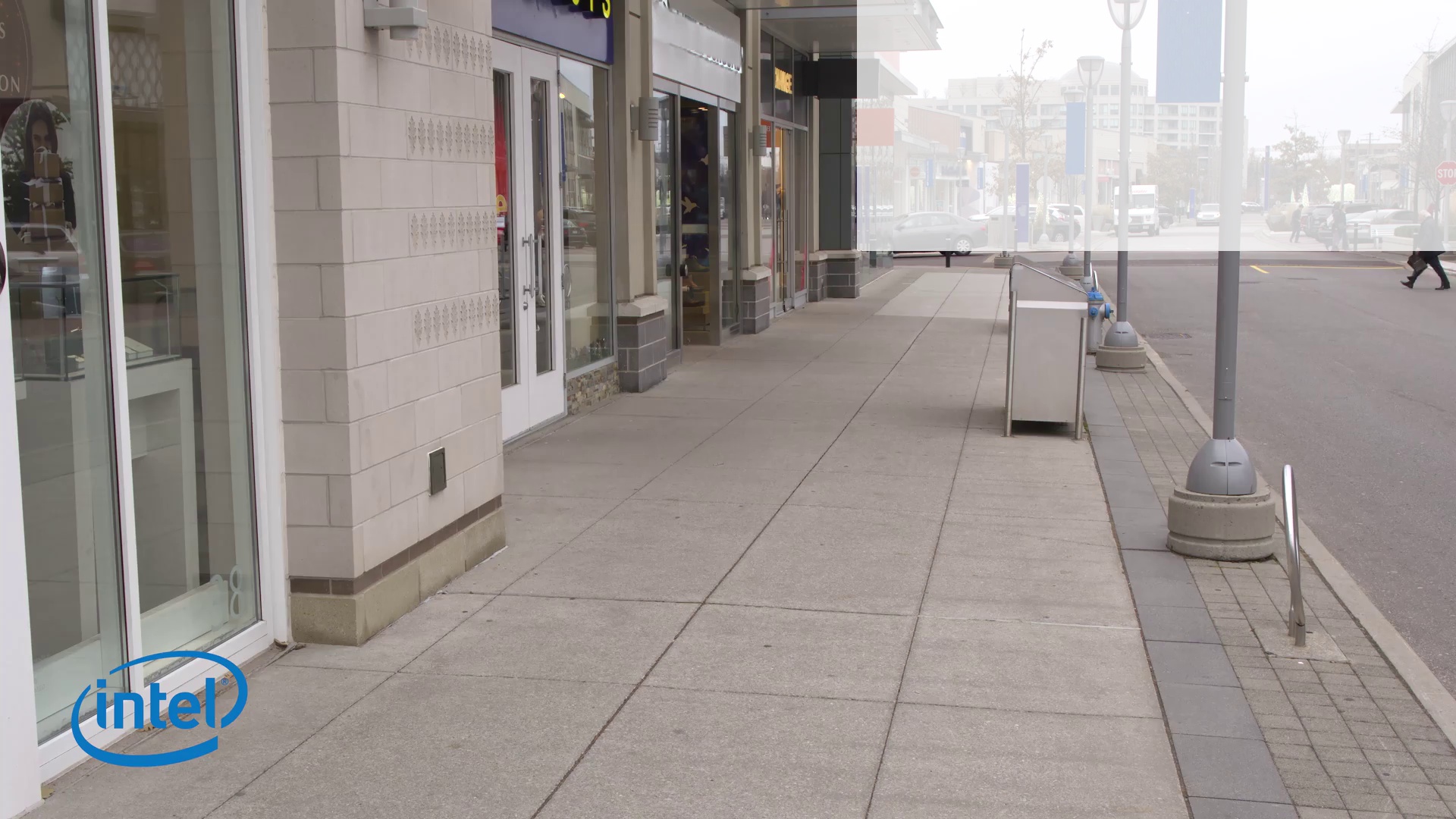} &
        \includegraphics[width=0.13\textwidth]{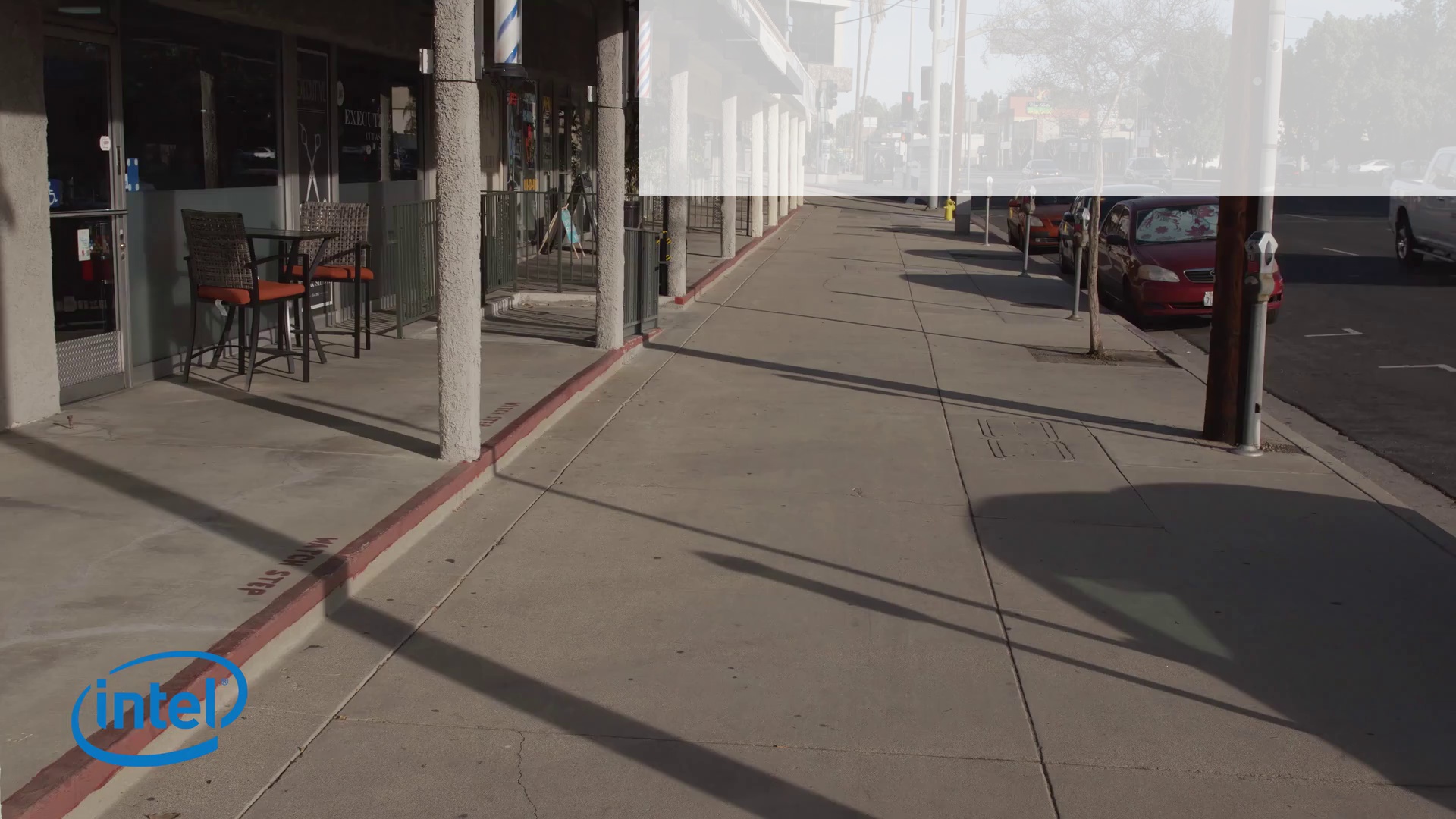} \\
        Airport-1 & Airport-2 & Airport-3 & Airport-4 & Mall-1/2 & Mall-3/4 & Pedestrian-1  \\
         \includegraphics[width=0.13\textwidth]{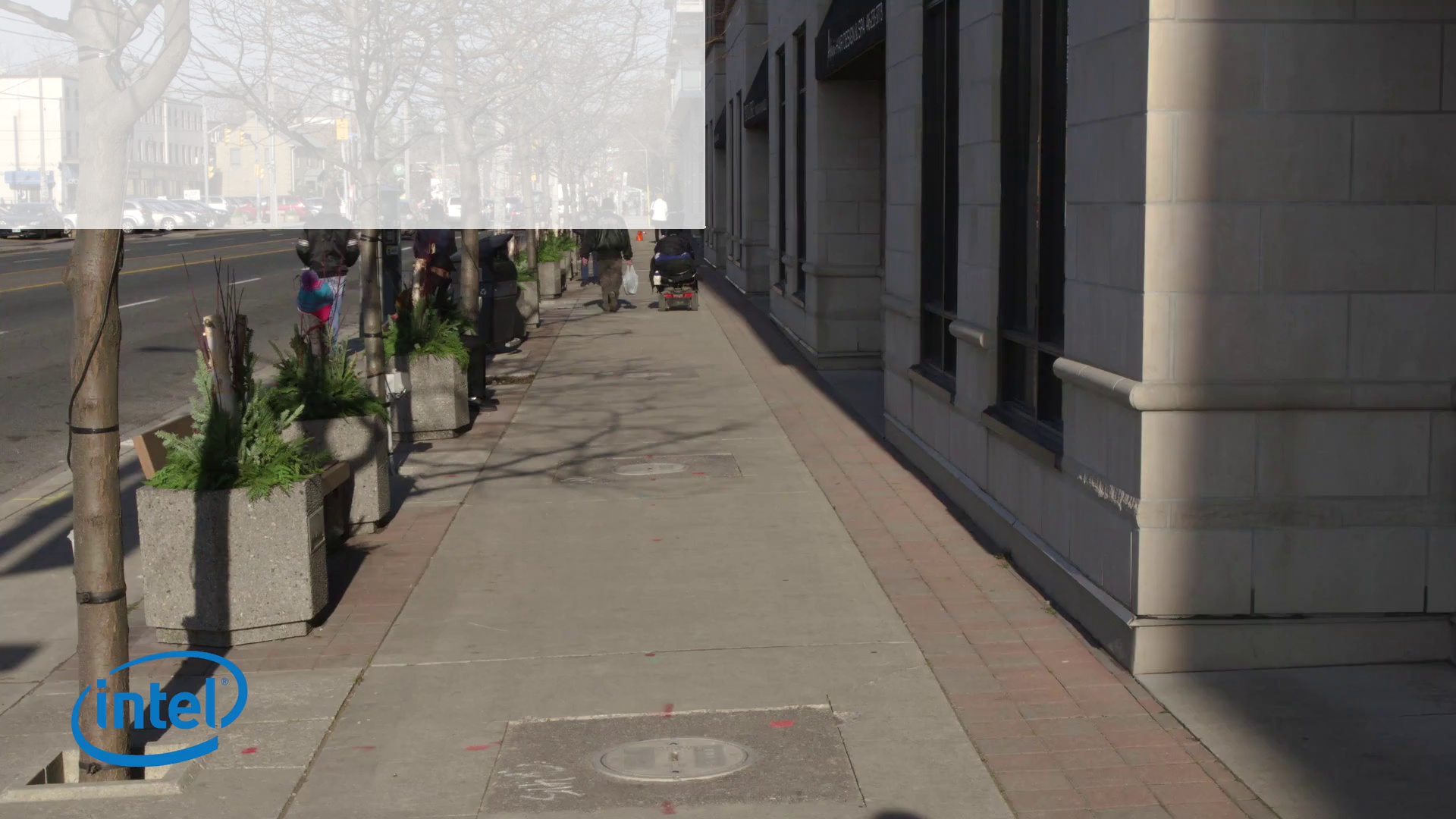} &
        \includegraphics[width=0.13\textwidth]{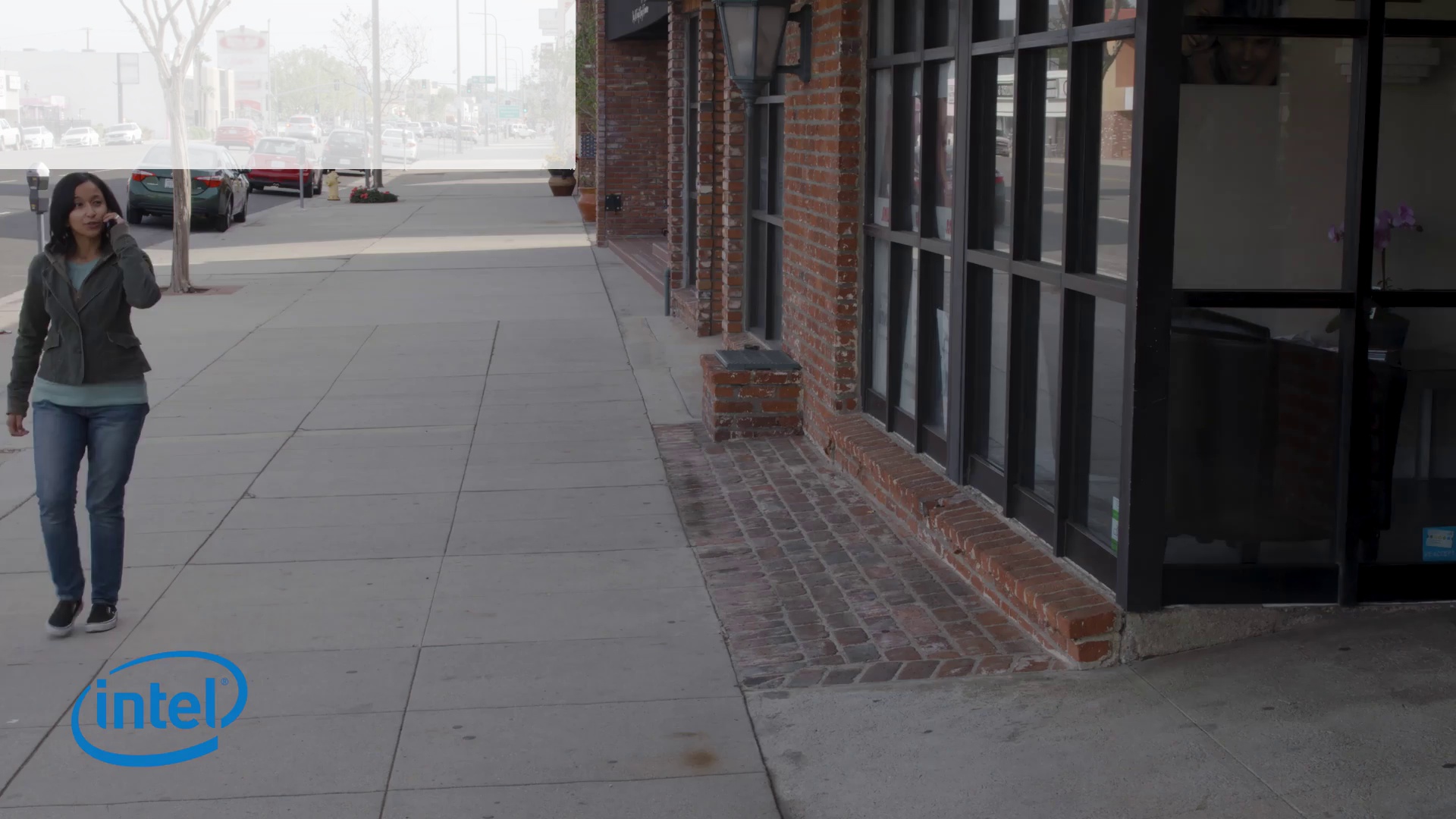} &
        \includegraphics[width=0.13\textwidth]{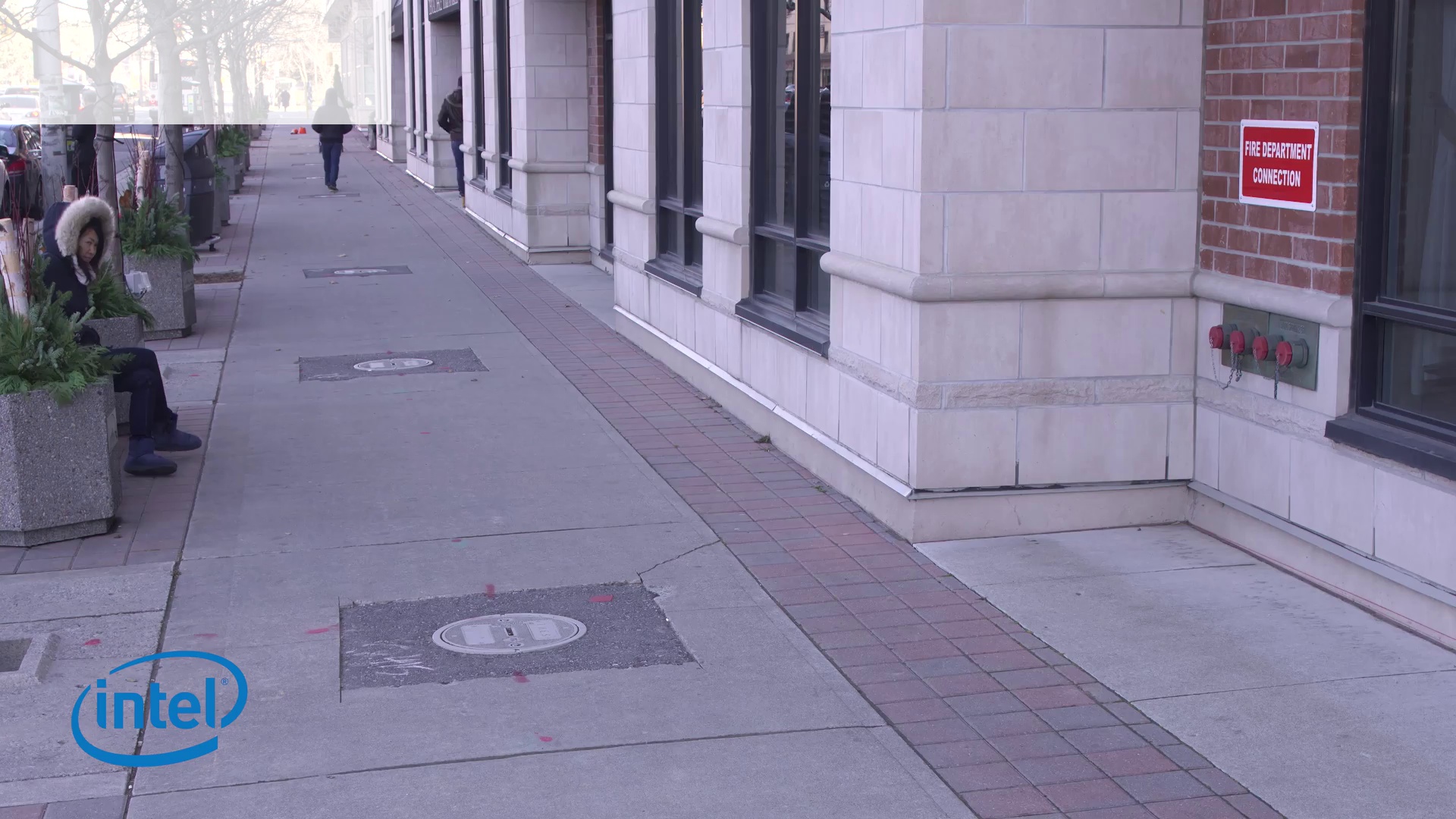} &
        \includegraphics[width=0.13\textwidth]{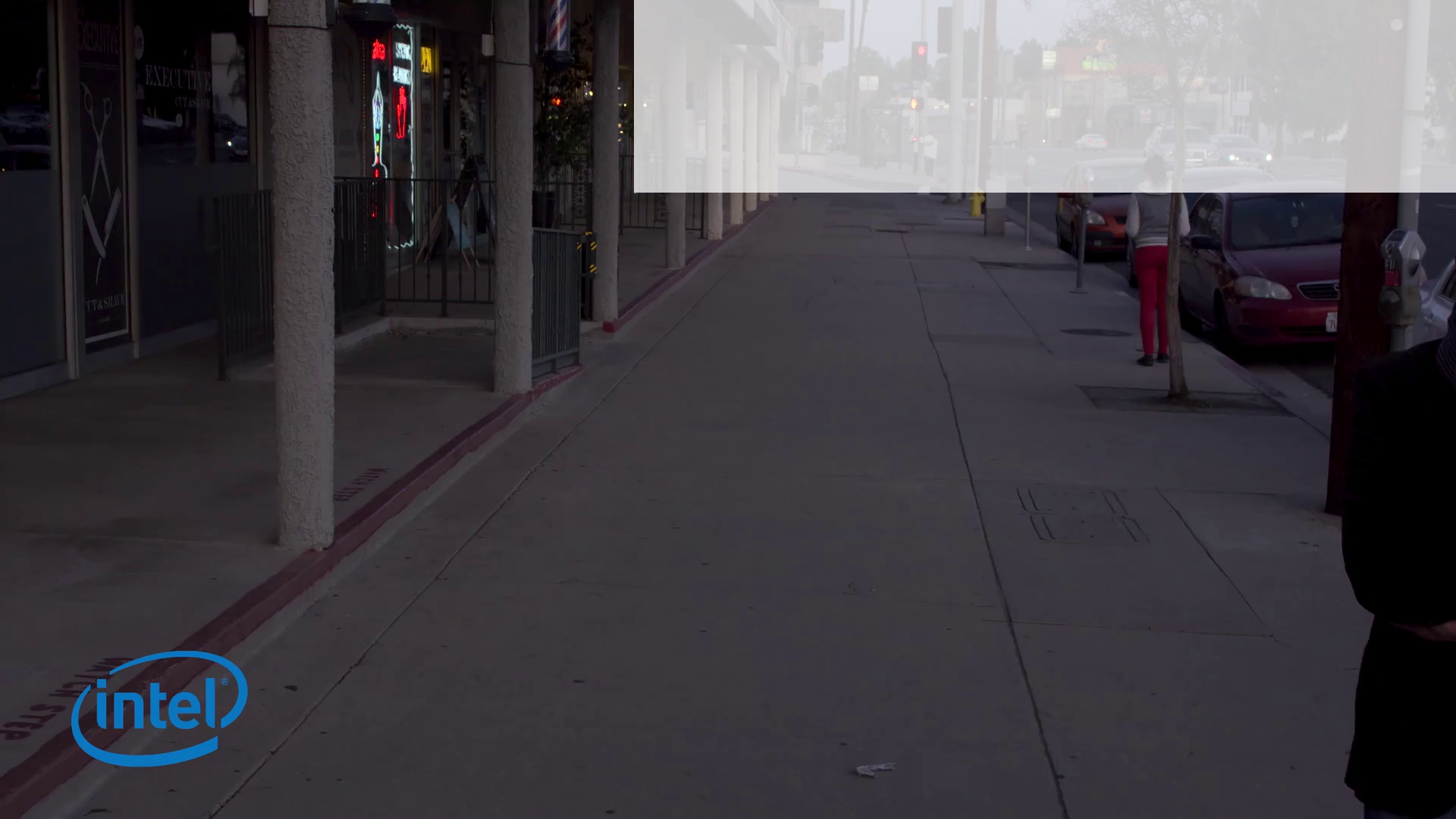} & 
        \includegraphics[width=0.13\textwidth]{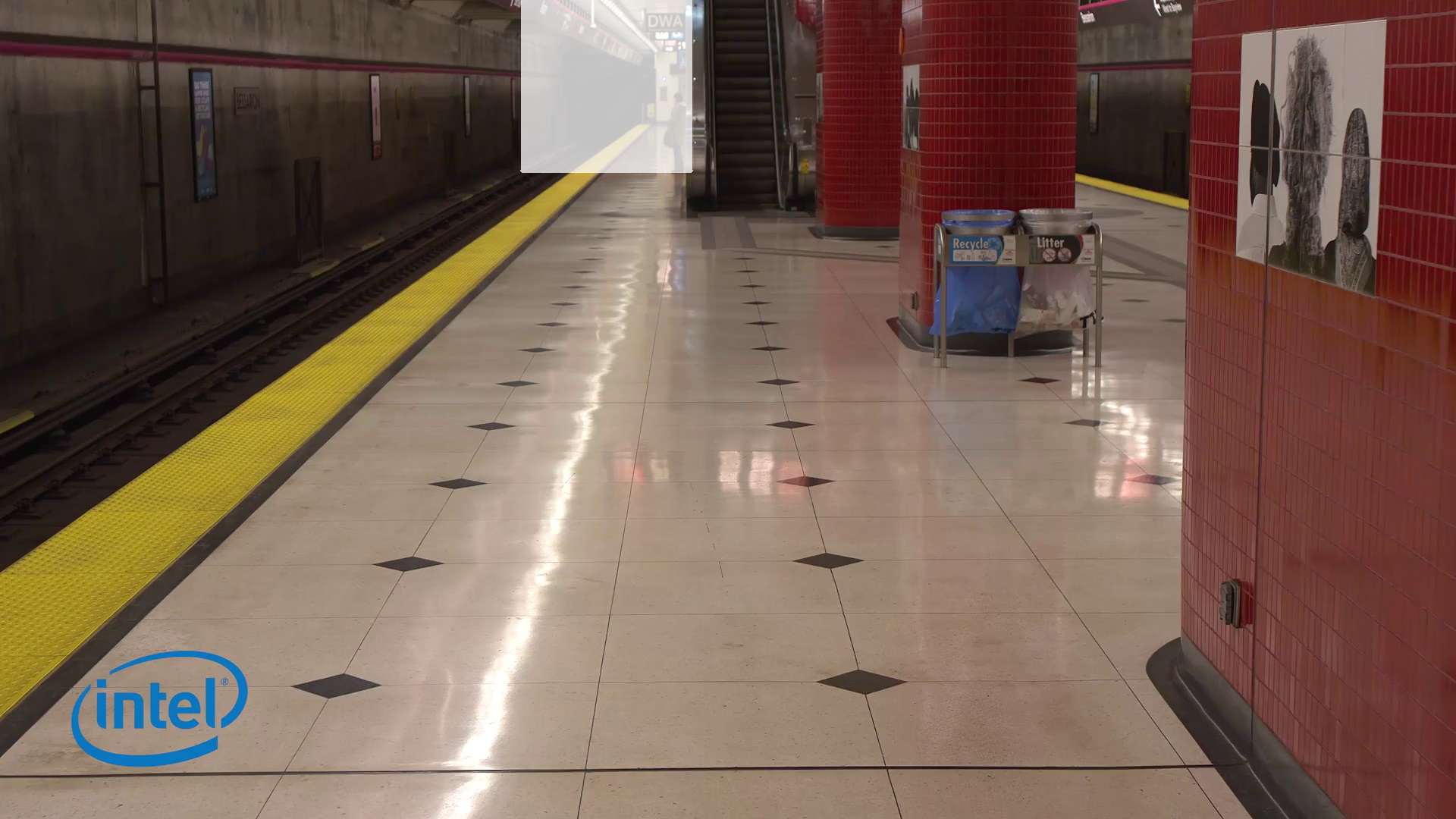} &
         \includegraphics[width=0.13\textwidth]{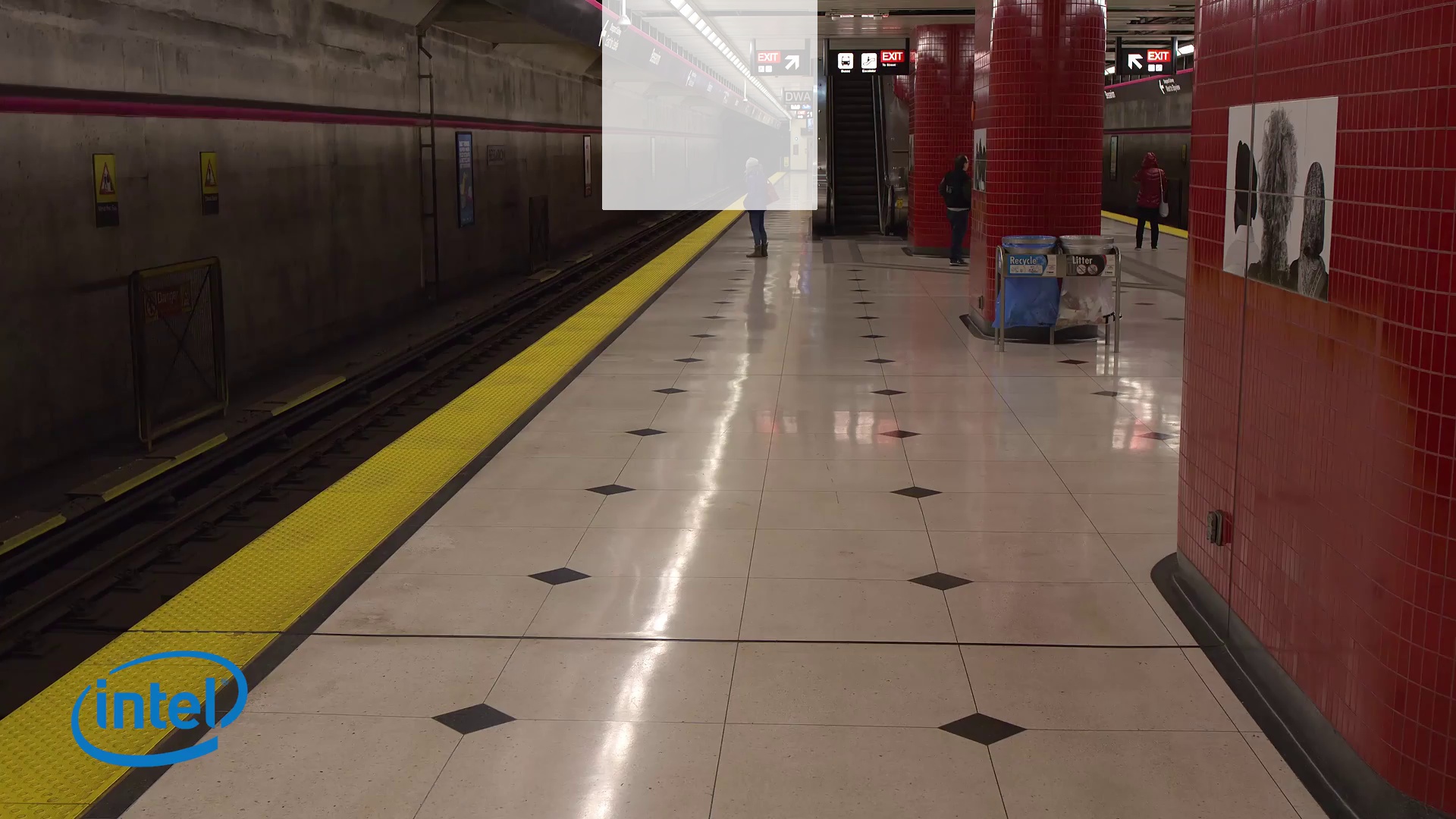} & 
        \includegraphics[width=0.13\textwidth]{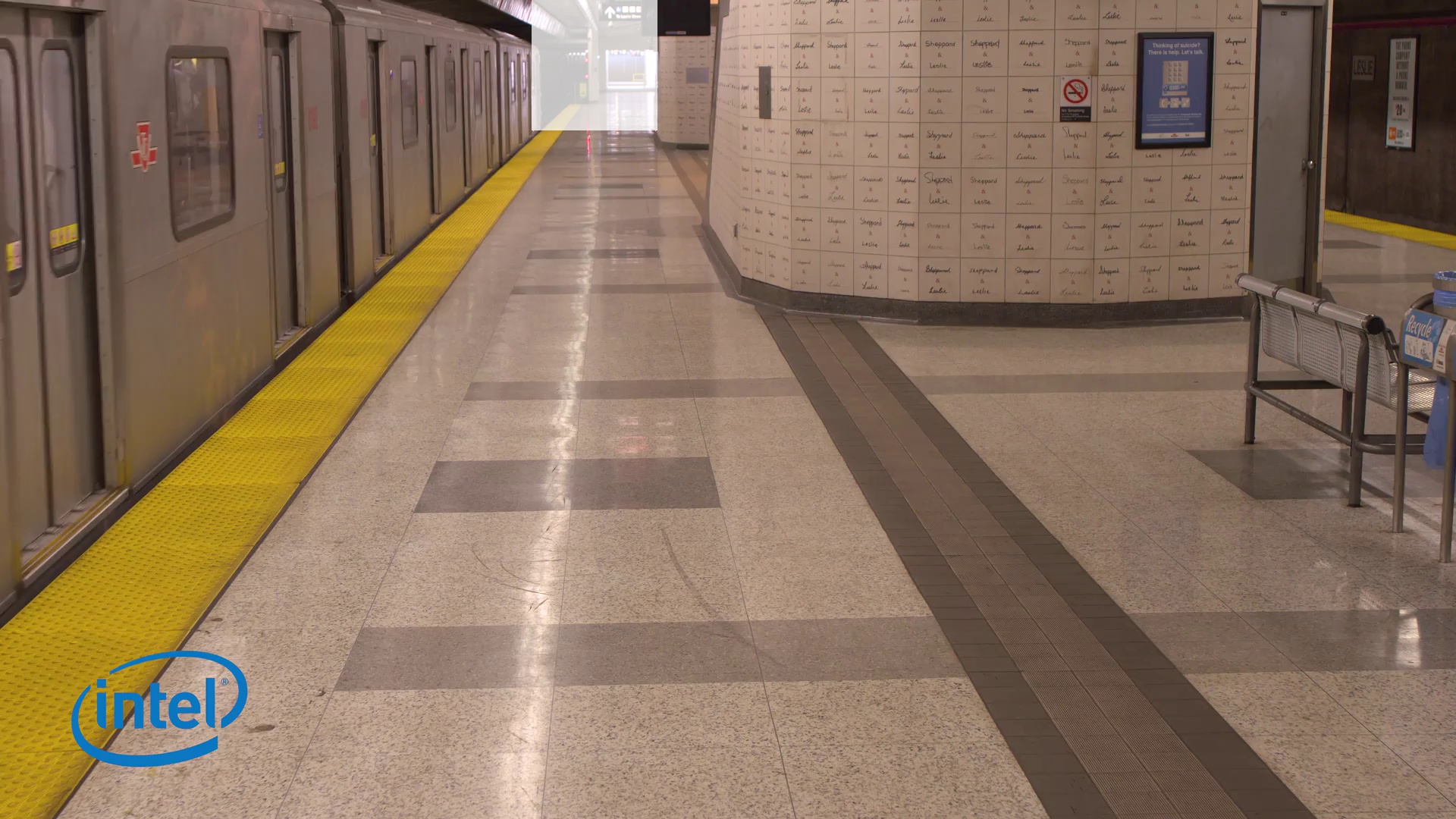} \\
        Pedestrian-2 & Pedestrian-3 & Pedestrian-4 & Pedestrian-5 & Subway-1 & Subway-2 & Subway-3  \\
    \end{tabular}
    \caption{First frame of each video, with the white shading indicating \textit{ignore areas} where annotations and estimations are not considered for the computation of the performance measures.}
    \label{fig:dataset_people}
\end{figure*}

\begin{figure*}[!t]
    \centering
    \includegraphics[trim=0 0 725 0, clip, width=0.9\columnwidth]{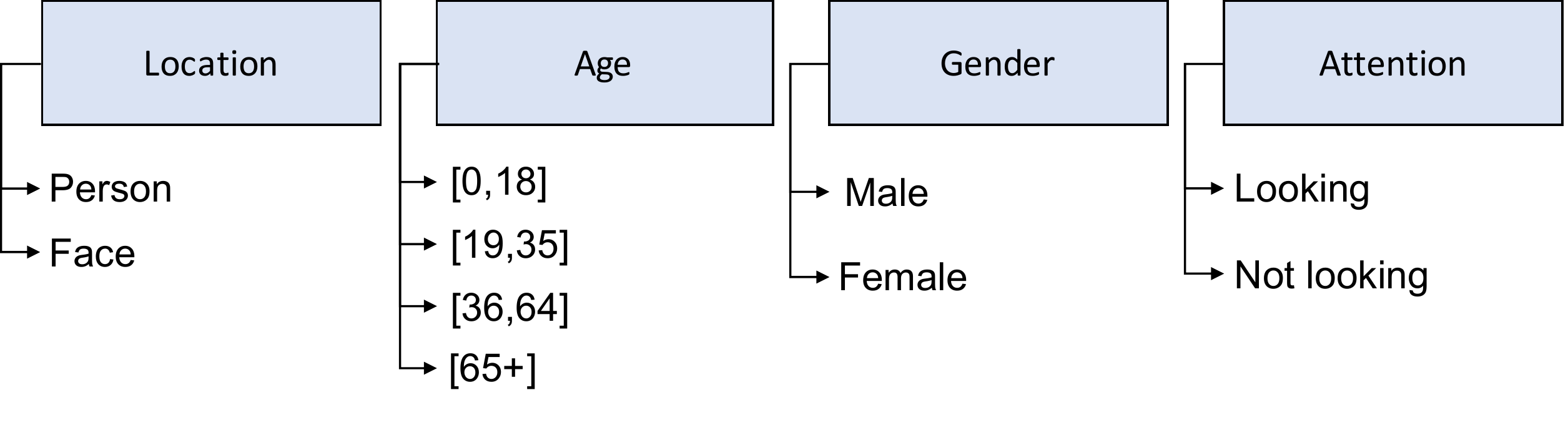}
    \includegraphics[trim=725 0 0 0, clip, width=0.9\columnwidth]{annotations_people.pdf}
    \caption{Attributes used in the annotations of the dataset.}
    \label{fig:annotations}
\end{figure*}

We employ A1-4 and C1-2 to estimate the instantaneous and cumulative number of people with OTS. The instantaneous number of people at $t$ is estimated as the number of detections/tracks at the current time. The cumulative number of people between any two instants, $t_1$ and $t_2$, is estimated as the number of unique identities $j$ during the considered time segment. 
Detectors (A1-2) simply assign a new identity to each detection without considering temporal relationships, thus producing an overcount.
As trackers require detectors and a temporal association of detections mechanism to prevent multiple counts of the same person over time, in general trackers require more computational resources than detectors. 
Besides, trackers (A3-4) use person detectors (as opposed to face detectors), therefore trackers may not have skills to differentiate people with OTS from those who do not have OTS.

We make the codes and pre-trained model weights of all baseline algorithms available at the project website.
Without altering the core of the original algorithms, we integrate changes to ensure that every algorithm processes the videos and generates the outputs following the same procedure. The codes for all baseline algorithms have been modified to enable both GPU (NVIDIA inference) and CPU (OpenVINO\footnote{At submission time, some operations required by A4 (i.e.~\textit{ScatterND}) are not supported by OpenVINO; hence, no OpenVINO optimization is used with this algorithm.} inference) computation in PyTorch (Table~\ref{tab:algorithms_summary}).

\begin{table*}[!t]
    \centering
    \caption{The dataset.}
    \label{tab:dataset_people}
    \resizebox{\textwidth}{!}{
    \begin{tabular}{lrrrrrrrrr}
        \specialrule{1.2pt}{0.2pt}{1pt}
        \multicolumn{2}{c}{\textbf{Video}} & \multicolumn{2}{c}{\textbf{Length}} & \multicolumn{2}{c}{\textbf{Unique people*}} & \multicolumn{4}{c}{\textbf{Number of localization annotations}}\\
        \cmidrule(lr){1-2} \cmidrule(lr){3-4} \cmidrule(lr){5-6} \cmidrule(lr){7-10}
        \multicolumn{1}{c}{Name} & \multicolumn{1}{c}{Illumin. [Lux]} & Time [min:sec] & Frames & All & OTS & People & Per-frame & Faces & Per-frame  \\
        \specialrule{1.2pt}{0.2pt}{1pt}
        Airport-1    & 500 & 5:21 & 9629 & 37 & 29 & 22062 & 2.4 $\pm$ 1.0 & 12832 & 1.4 $\pm$ 1.1 \\ 
        Airport-2    & 500 & 5:34 & 10008 & 35 & 29 & 23600 & 2.7 $\pm$ 1.4 & 14214 & 1.6 $\pm$ 1.2 \\ 
        Airport-3    & 500 & 6:26 & 11578 & 47 & 44 & 26704 & 2.4 $\pm$ 1.2 & 17849 & 1.6 $\pm$ 1.0 \\ 
        Airport-4    & 500 & 5:08 & 9247 & 61 & 56 & 43685 & 4.7 $\pm$ 2.0 & 17792 & 1.9 $\pm$ 1.2 \\  \hline
        Mall-1       & 300 & 4:38 & 8344 & 158 & 111 & 106852 & 12.8 $\pm$ 2.3 & 45835 & 5.5 $\pm$ 1.8 \\ 
        Mall-2       & 300 & 3:41 & 6626 & 145 & 105 & 95417 & 14.4 $\pm$ 3.7 & 42779 & 6.5 $\pm$ 2.3 \\ 
        Mall-3       & 800 & 5:25 & 9740 & 33 & 30 & 37120 & 3.8 $\pm$ 1.5 & 18906 & 1.9 $\pm$ 1.2 \\ 
        Mall-4       & 800 & 6:04 & 10931 & 53 & 50 & 47113 & 4.3 $\pm$ 1.6 & 32038 & 2.9 $\pm$ 1.3 \\   \hline
        Pedestrian-1 & 60000 & 5:40 & 10202 & 18 & 17 & 39680 & 4.0 $\pm$ 1.7 & 19859 & 2.0 $\pm$ 1.4 \\  
        Pedestrian-2 & 40000 & 6:15 & 11262 & 56 & 40 & 58477 & 5.2 $\pm$ 1.7 & 25042 & 2.2 $\pm$ 1.6 \\
        Pedestrian-3 & 7000 & 5:41 & 10220 & 27 & 25 & 22738 & 2.3 $\pm$ 1.2 & 13915 & 1.4 $\pm$ 1.0 \\ 
        Pedestrian-4 & 5500 & 4:32 & 8166 & 27 & 25 & 33031 & 4.0 $\pm$ 1.4 & 16248 & 2.0 $\pm$ 1.0 \\ 
        Pedestrian-5 & 250 & 2:58 & 5350 & 11 & 11 & 24476 & 4.6 $\pm$ 1.6 & 13504 & 2.5 $\pm$ 1.8 \\  \hline
        Subway-1     & 180 & 3:13 & 5795 & 17 & 17 & 36828 & 6.5 $\pm$ 3.1 & 25884 & 4.6 $\pm$ 2.7 \\
        Subway-2     & 180 & 2:32 & 4549 & 29 & 28 & 45125 & 9.9 $\pm$ 2.8 & 24248 & 5.3 $\pm$ 2.4 \\
        Subway-3     & 200 & 5:45 & 10342 & 31 & 29 & 85358 & 8.5 $\pm$ 2.9 & 35460 & 3.6 $\pm$ 1.7 \\
        \specialrule{1.2pt}{0.2pt}{1pt}
        Overall & [180,60000] & 78:53 & 141989 & 785 & 646 & 748266 & 5.4 $\pm$ 3.9 & 376405 & 2.7 $\pm$ 2.1 \\ 
        \specialrule{1.2pt}{0.2pt}{1pt}
        \multicolumn{10}{l}{*People re-entering the field of view, after exiting it for longer than 10~seconds, are considered as a new (unique) person.}
    \end{tabular}
    }
\end{table*}

\section{Dataset}
\label{sec:dataset}

\subsection{Videos}
The dataset was collected in settings that mimic real-world signage-camera setups used for AVA.
The dataset is composed of 16 videos recorded at different locations such as airports, malls, subway stations, and pedestrian areas. Outdoor videos are recorded at different times of the day such as morning, afternoon, and evening.
The dataset is recorded with Internet Protocol or USB fixed cameras with wide and narrow lenses to mimic real-world use cases. Videos are recorded at 1920 $\times$ 1080 resolution and 30~fps.
The dataset includes videos of duration between 2 minutes and 30 seconds, and 6 minutes and 26 seconds, totaling over 78 minutes, with over 141,000 frames.
The videos feature 34 professional actors with multiple ethnicities, with ages from 10 to 80, and including male and female genders. People have been recorded with varied emotions while looking at the signage. 
A sample frame of each location is shown in Figure~\ref{fig:dataset_people}. {We show sample frames with a reduced number of people for facilitating the visualization of the background.} For the mall location, two videos are at different times: indoors (Mall-1/2) and outdoors (Mall-3/4).

\subsection{Annotations}
A professional team of annotators used the Intel Computer Vision Annotation Tool~\cite{annotators} {from OpenVINO ecosystem} to fully annotate all videos with the following attributes (Figure~\ref{fig:annotations}): bounding boxes for face and body of the people, identity, age, gender, attention, pose, orientation, and occlusions.
{Annotations were generated for every key-frame. The key-frames were selected depending on the behavior and the location with respect to the camera of the person to be annotated. People closer to the camera (or moving faster) were annotated more often than people that are farther away (or moving slower). Inter-key-frames annotations were generated using linear interpolation. The annotations were validated by expert annotators that checked the consistency of person-face groups, person identity within a video and across videos, and an individual visual inspection of each of the annotated attributes.} For preventing the analytics to focus on very small (far from signage) people, who are likely to not have OTS, and to simplify the annotation process, we define a region in some scenarios where people are omitted, and thus not annotated. We refer to these regions as \textit{ignore area}, shown as a white shading in Figure~\ref{fig:dataset_people}. Estimations within the ignore areas are also omitted.
The annotations maintain the identity of each person throughout the same video, even if the person exits and re-enters into the field of view, and even across videos. However, for the purpose of AVA benchmarking, when an actor exits and re-enters into the field of view of a camera within the same video after more than 10 seconds, we consider the same actor to have a new identity.
Each video includes a range between 11 and 158 \textit{unique} people. The dataset annotation includes a total of 785 unique people, and over a million annotated bounding boxes. Most of the people present in the dataset have OTS and roughly 10\% of them looked directly at the camera. The main characteristics of the dataset are summarized in Table~\ref{tab:dataset_people}. 

\section{Benchmark - Results and discussion }
\label{sec:benchmark}

\begin{table}[!t]
    \centering
    \setlength{\tabcolsep}{0pt}
    \caption{Main characteristics of the systems (S) used in the experiments. KEY -- inf.: inference; NV.: NVIDIA.
    }
    \label{fig:systems}
    \resizebox{\columnwidth}{!}{%
    \begin{tabular}{lccccccc}
    \specialrule{1.2pt}{0.2pt}{1pt}
        \textbf{Short name} & \textbf{S1} & \textbf{S2} & \textbf{S3} & \textbf{S4} &  &  &  \\
        & \includegraphics[height=1cm]{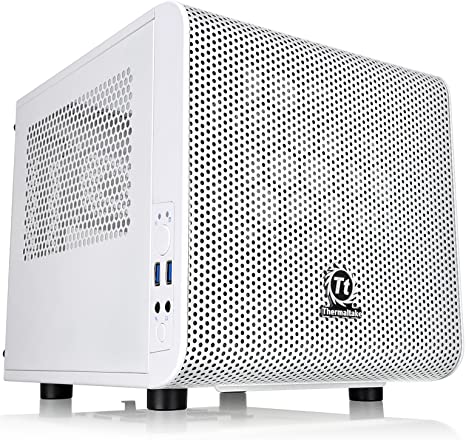} & \includegraphics[height=1cm]{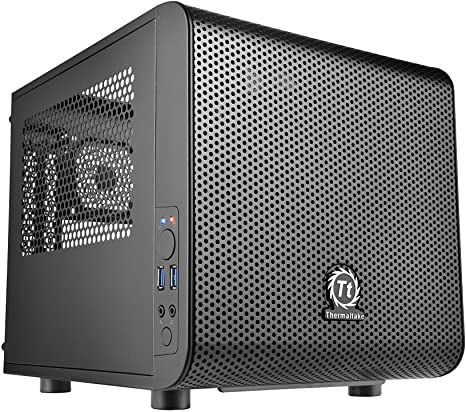} & \includegraphics[height=1cm,angle=-90,origin=c]{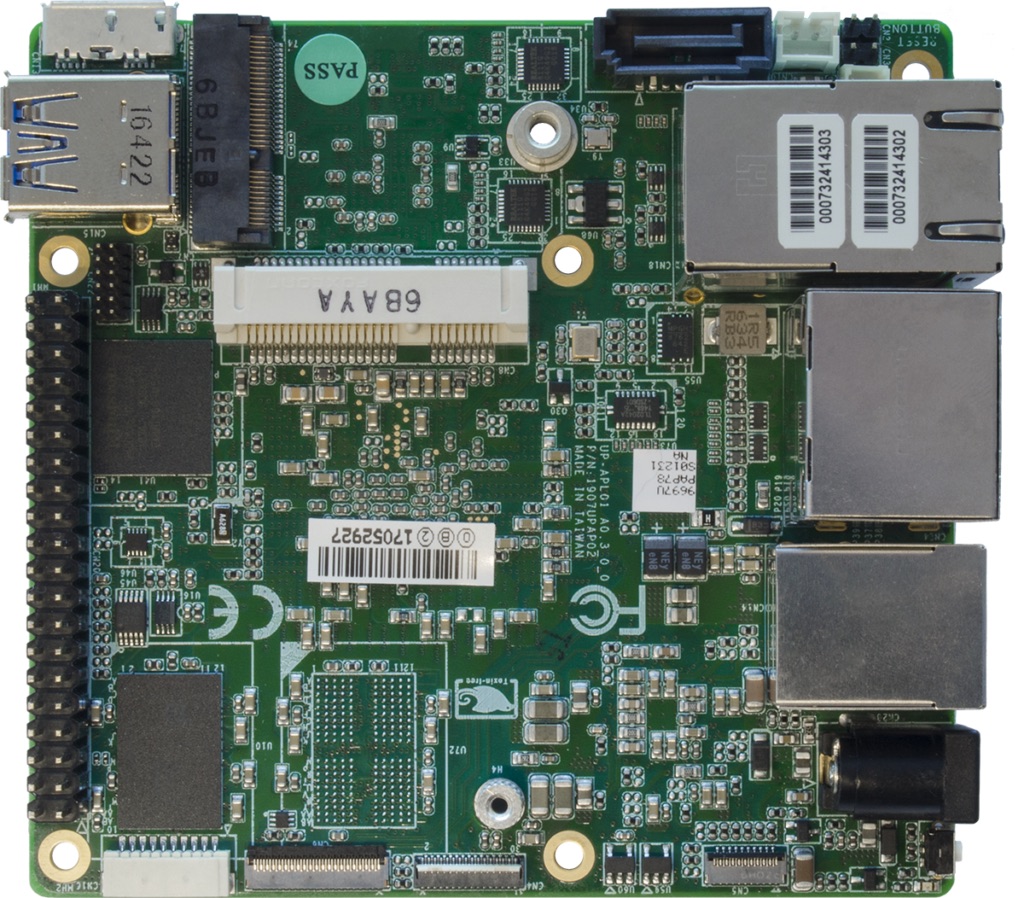} & \includegraphics[height=1cm]{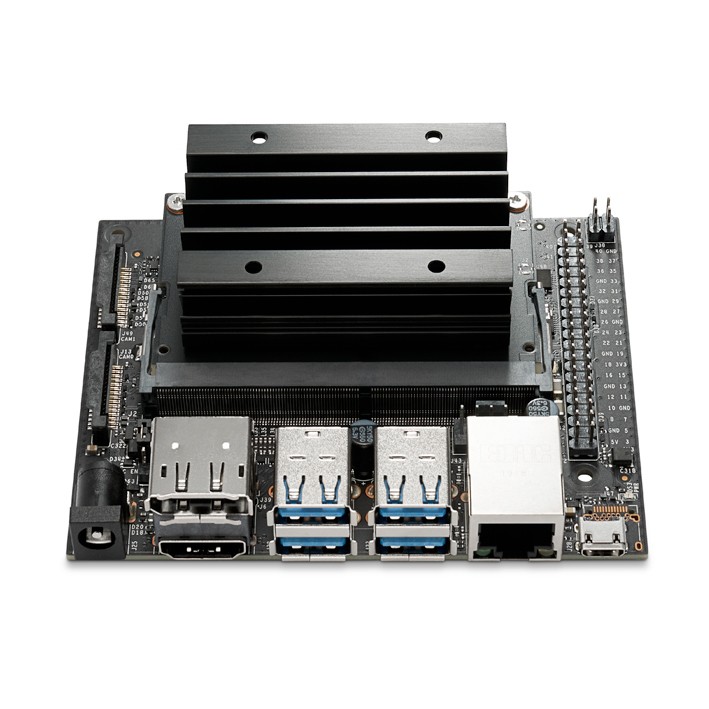} &  &  &  \\
        CPU & Intel i5 & AMD R.5 & Intel Atom x7 & ARM &  &  &  \\
        GPU & GTX1660Ti & GTX1660Ti & Intel Gen9 HD & NV. Pascal &  &  &  \\
        Memory [GB] & 32 & 32 & 8 & 8 &  &  &  \\
        Storage [GB] & 512 & 512 & 64 & 32 &  &  &  \\
        Consump. [W] & 450 & 450 & 18 & 7.5 &  &  &  \\
        NVIDIA inf. & \checkmark & \checkmark &  & \checkmark &  &  &  \\
        OpenVINO inf. & \checkmark &  & \checkmark &  &  &  & \\
        \specialrule{1.2pt}{0.2pt}{1pt}
    \end{tabular}
    }
\end{table}

\begin{table}[!t]
    \centering
    \caption{Input frame rate for each localization algorithm and system when using CPU (GPU) inference. The selected frame rate allows the algorithm-system combination to achieve (near) real-time processing speed.}
    \label{tab:algorithms_frame_rates}
    \resizebox{\columnwidth}{!}{%
    \begin{tabular}{ccccc}
     \specialrule{1.2pt}{0.2pt}{1pt}
         \textbf{Algorithm}  & \textbf{System 1}    & \textbf{System 2}   & \textbf{System 3}   & \textbf{System 4} \\
     \specialrule{1.2pt}{0.2pt}{1pt}
         A1         & 30 (30)     & 3 (30)     & 3          & (3)     \\
         A2         & 30 (30)     & 3 (30)     & 3          & (3)     \\
         A3         & 30 (30)     & 3 (30)     & 3          & (3)     \\
         A4         & 30 (30)     & 1 (30)     & 1          & (1)     \\
         A5-6       & 30 (30)     & 3 (30)     & 3          & (3)     \\
    \specialrule{1.2pt}{0.2pt}{1pt}
    \end{tabular}
    }
\end{table}

\input{counting}
\input{droping}
\input{age_gender}

\subsection{Experimental setup}

We evaluate the performance of the algorithms on four systems that enable on-the-edge AVA processing. The systems' properties are summarized in Table~\ref{fig:systems}.
The algorithms are executed in both GPU and CPU (one core), separately. 
GPU inference is used for systems with an integrated NVIDIA GPU.
CPU inference is used for all systems, and it can be native (i.e.~without optimization) or optimized using OpenVINO. The use of OpenVINO optimization depends on the system and the algorithm. Systems must be equipped with an Intel processor, and algorithms must be compatible with the OpenVINO optimization. For the baseline algorithms (A1-6), all algorithms but A4 are OpenVINO compatible.

We define as real-time processing the capability of a system-algorithm pair to complete the analytics for $I_t$ before the new data frame $I_{t+1}$ is available.
When a system-algorithm pair does not achieve real-time processing (e.g.~input data is at 30~fps and the processing speed is at 1~fps), one can reduce the frame rate and/or resolution of the videos. We decide to maintain the resolution of the videos and intentionally reduce the frame rate (i.e.~drop frames) of the input videos to ensure that every system-algorithm pair performs (near) real time. Table~\ref{tab:algorithms_frame_rates} shows the input frame rate of the videos that we use for all experiments for every system-algorithm pair.
For System 1, all algorithms run with 30~fps videos with both GPU and CPU inference.
For System 2, all algorithms run with 30~fps videos with GPU and 3~fps videos with CPU, except for A4 with CPU that runs with 1~fps videos.
For Systems 3 and 4, all algorithms run with 3~fps videos, except for A4 that runs with 1~fps videos.

Most of the results shown next are represented by box plots. The horizontal line within the box shows the median; the lower and upper edges of the box are the 25-percentile and 75-percentile; and, the bottom and top edges show the minimum and maximum values.

We provide an \textit{online evaluation tool} that allows one to effortlessly assess the performance of AVA algorithms in the proposed dataset. Further information regarding the requested data format and use of the evaluation tool is available at the project website.


\subsection{Localization}
\label{sec:results_localization}

\input{localization}

Figure~\ref{fig:localization} shows the evaluation results of A1-2 on face localization, and A3-4 on person localization. 
Regarding the detectors, A2 obtains a higher precision but lower recall than A1. This indicates that A1 generates a larger amount of false positives than A2. However, the median results for F1-Score show that A1 outperforms A2 by a small amount.
Regarding the trackers, A3-4 obtain comparable results across the systems.
For instance, in System 1 with GPU, A1 obtains a F1-Score 0.17 higher than A2, and A3 and A4 obtain a median F1-Score of 0.80 and 0.81, respectively. 
When considering the distance between people-signage, results show that all algorithms are able to localize a larger amount of people (i.e.~higher recall) when people are closer to the camera.
For instance in System 1 with GPU,  A1 obtains a median recall of 0.94/0.68 for closer/farther faces; and A3-4 trackers obtain a median recall of 0.91 for closer people, and 0.55 and 0.63 for farther people, respectively.
When analyzing the performance measures as a function of the occlusion levels, results indicate that the recall drops when faces/bodies are partially or heavily occluded.
For instance in System 1 with GPU, while the median recall for \textit{non-occluded} people and algorithms A1, A3, and A4 is above 0.8; the median recall drops to values below 0.6 for \textit{partial occlusions} and below 0.5 for \textit{heavy occlusions}.

\subsection{Counting}
\label{sec:results_people_counting}

Figure~\ref{fig:count} shows the counting evaluation in terms of MOE, MPE, and COE.
Regarding \textit{Mean Opportunity Error} (MOE), 
A1-2 (i.e.~detectors) have errors up to 10 people for some videos considering all systems. Detector obtains a large range of errors, indicating a non-uniform performance throughout different videos of the dataset.
On the contrary, A3-4 (i.e.~trackers) have a smaller median error as well as a smaller range of errors, which are always under 4 people. 
As observed in localization, algorithms commit fewer errors with people who are closer to the camera/signage, than with people who are farther.
Algorithms that consider the body of people, instead of their faces (i.e.~A3-4) obtain a lower \textit{Mean People Error} (MPE). This is expected, as these tracking algorithms have no skill in determining the OTS of people; thus, when considering all people regardless of their OTS, a lower error is obtained. Face detection algorithms (A1-2) obtain median MPE above 2 people, whereas body person tracking algorithms (A3-4) obtain median MPE below 2 people.
Similar performance is obtained across systems and inference units (i.e.~GPU vs CPU) except for A1, which has a smaller error range when executed in CPU.

When the task is to count the cumulative number of unique people with OTS, \textit{Cumulative Opportunity Error} (COE) indicates that tracking algorithms obtain more accurate count than detection algorithms. With System 1, while the tracking algorithm A3-4 obtains a median COE under 1.02 and 4.20, respectively; detection algorithms A1-2 obtain a median COE of at least two orders of magnitude higher.
In this case, algorithms that are using the GPU for inferring obtain in general higher error than when using CPU. This is due to the fact that algorithms that process more frames (i.e.~GPU) are more prone to overcount the cumulative number of people than when processing fewer frames (i.e.~CPU). Algorithms using CPU inference use as input videos with lower frame rate than when using GPU inference (e.g.~System 2 uses 3~fps videos with CPU inference and 30~fps videos with GPU inference), thus the overcount is likely to be reduced as the number of processed frames is reduced. This effect can also be seen in the results with System 1, where CPU and GPU use the same video frame rate, therefore, obtain very similar results regardless of the inference type.
When the task is to estimate the cumulative number of people over a specific segment of time, \textit{Temporal Cumulative Opportunity Error} (TCOE) results show that the error increases monotonically when the duration of the segment ($D$) increases (Figure~\ref{fig:count_TCOE}). Also, it can be observed that detectors (A1-2) obtain several orders of magnitude higher TCOE than trackers (A3-4). With System 1 GPU, trackers obtain TCOE of 1.5-19 (A3) and 6-70 (A4). The most accurate algorithm for this task is A3, which can estimate the cumulative number of people with OTS for 10 (120)-second segment video with a median TCOE of $\pm$2.20 ($\pm$18.81), with System 1 and GPU.

\textit{Error vs input frame rate.}
\input{commercial}
\input{speed}
\input{pervideo}

Results indicate that the input frame rate of the videos has an important effect on the performance of the algorithms. Thus, we analyze the trade-off between the error obtained by the algorithms and the input frame rate of the videos. This analysis is divided into two experiments. 
In the first experiment, we intentionally modify the input frame rate of the videos to \mbox{$\gamma=\{0.25,0.5,1,2,6,7.5,10,15,30\}$}~fps, and then, we compare the performance of the localization algorithms in terms of MOE, MPE, and COE. Figure~\ref{fig:count_dropping} shows the results. For MOE and MPE, in the first row, the errors are mostly flat.
This indicates that those performance measures are not affected by the input frame rate of the videos. This is expected as these two measures are performed on a frame-by-frame basis. Therefore, when reducing the frame rate of the input videos, the number of frames that the algorithms compute is reduced but this reduction does not affect to the performance of the algorithms, as non-seen frames are used neither for inference nor evaluation.
However, when computing cumulative counts across time, the performance of the algorithms can be affected. In fact, the second row of Figure~\ref{fig:count_dropping}, we observe that COE varies for A3-4 when the input frame rate of the videos change. 
COE monotonically increases when the frame rate increases for detection algorithms (A1-2).
The performance obtained by A3 does not change substantially for different input frame rate, and its minimum occurs around 6~fps. Surprisingly, COE monotonically increases for A4 when the frame rate increases. However, when considering absolute values, the error is considerably lower than the one obtained by detection algorithms. This indicates that higher frame rates do not necessarily ensure higher counting performance.
In the second experiment, we apply the same frame rate reduction than in the previous experiment, and we compute TCOE for different segment duration of $D=\{10,20,30,60,90,120\} \, \gamma$ frames (i.e.~10, 20, 30, 60, 90, 120~seconds). The results are in Figure~\ref{fig:count_dropping_TCOE}. Results in the first two rows indicate that TCOE monotonically increases for detection algorithms (A1-2) when the frame rate increases, and also when the duration of the segment increases. As mentioned before, this occurs as detectors have no skill for counting the cumulative number of people. When the segment increases its duration, the overcount accumulates further, thus TCOE also increases.
Unlike detectors, trackers (A3-4), shown in the last two rows of Figure~\ref{fig:count_dropping_TCOE}, obtain a more stable TCOE for any segment duration and frame rate. Segments with larger duration produce a slight increment in the TCOE. Note that as A4 cannot be optimized using OpenVINO for CPU computation, we limit the input frame rate of the videos up to 1~fps as higher frame-rate videos make the algorithm very slow. In this experiment, A3 is the most accurate baseline algorithm under comparison.
Visual and detailed per-video results obtained by each algorithm are available at the project website.

\subsection{Age and gender estimation}

Age and gender estimation algorithms require as input a crop with the person's face. We employ the face detector that obtains the highest recall (A1) as detector. To fairly compare the age and gender estimation algorithms and disregard the localization performance, we compute the age for correctly detected (i.e.~true positive) faces only.
The results with A5-6 for age estimation are shown in Figure~\ref{fig:age_gender_results} (first three columns), and for gender estimation Figure~\ref{fig:age_gender_results} (last three columns).

Age estimation results indicate that this task is challenging as the highest median F1-Score obtained is below 0.7. 
While the highest results are obtained for classes [19,34] and [35,65] with median F1-Score between 0.3 and 0.7, even lower F1-Scores are obtained for [0,18] and [65+] classes where algorithms obtain median F1-Score under 0.1. This suggests that the baselines algorithms, A5-6, are not skilled in determining the age of younger and older people, where both algorithms obtain a very low recall for these classes.
Similar results are obtained regardless of the system and inference type (i.e.~CPU and GPU). 

Gender estimation results indicate that A5-6 algorithms obtain a similar F1-Score performance for both classes (\textit{male}, \textit{female}). For the \textit{male} class, both algorithms obtain a precision between 0.4 and 0.8 with a recall between 0.6 and 1.0. The opposite behavior happens for the class \textit{female}, obtaining higher precision than recall.
For this task, results indicate that all systems and inference types behave similarly. Note that A6 is not able to run in System 4 due to memory limitations.

\subsection{Commercial solutions}

As we did not have access to the source codes, the commercial solutions (C1-2) are executed in two external systems equipped with an Intel i7 CPU. To preserve the integrity of the complete dataset, until submission time, we only shared a subset of the dataset with the creators of C1-2. Thus, for this comparison, we use a subset of the dataset that is composed of the following videos \textit{Airport-1}, \textit{Airport-2}, \textit{Mall-3}, \textit{Mall-4}, \textit{Pedestrian-2}, and \textit{Pedestrian-3}.
As a reference, we show next the results obtained by the baseline algorithms (A1-6) with System 1 GPU in the same subset of videos.

The results of this experiment are reported in Figure~\ref{fig:results_commercial}.
For \textit{localization}, A1-2 and C1-2 are evaluated for face localization, and A3-4 are evaluated for person localization.
Results show that both commercial solutions (C1-2) perform similarly than A2 for all performance measures, while A1 outperforms other face detectors obtaining a recall higher than 0.4 and similar precision.
Person detectors (A3-4) obtain higher detection performance than face detectors with F1-Score around 0.8 and precision over 0.85.
\textit{Count} results show that all baselines and the commercial solutions obtain very similar MOE with median errors around 1 and always under 2.5, except for A1 that obtains a median MOE of 3.5 with values up to 8.
Commercial solutions obtain similar MPE errors than A2 with a median error of around 2 people. A3-4 obtain lower MPE indicating that these algorithms obtain a lower error (below 1 in its median) when counting instantaneous people regardless of their OTS. A1 obtains a higher error than other algorithms with median MPE around 4. 
Regarding COE, the commercial solution C1 obtains the lowest error with a median of only 0.20 and a small range of errors. This indicates that the algorithm can count the cumulative number of people in each complete video with an error below $\pm$ 1 person.
The authors of C2 have a different definition of OTS than the one described in this paper. C2 define OTS as any person within the field of view, regardless of their face visibility. Therefore, we also compute, for C2 only, the CPE, where all people visible on the field of view of the video are considered to have OTS. The results are on the last COE bar.

When we consider the ability of the algorithms for counting the cumulative number of people for different segment duration, algorithms obtain an average TCOE between 1 and 50 (Figure~\ref{fig:count_commercial_TCOE}). C1 is again the algorithm that obtains the lowest error for all segment duration, achieving TCOE between 2 and 4.

In addition to showing the results using the proposed performance measures, we show the raw estimated and annotated counts for a set of selected videos in Figure~\ref{fig:perframe}. 
We can observe that for instantaneous count most of the algorithms often estimate the number of people in the surrounding of the annotated value. An exception here is A1, which often generates multiple false positives, thus overestimating the actual number of people with OTS for a given instant.
Regarding the cumulative count (second column of Figure~\ref{fig:perframe}), and as done for COE, we show the estimated cumulative count of unique people with OTS from the initial frame. The charts indicate that the most accurate algorithms in this task are C1 and A3; and that the cumulative count is a challenging task and even state-of-the-art trackers (i.e.~A4) and commercial solutions (i.e.~C2) might drift from the actual number of people.

The detailed per-video results are available at the project website.

\subsection{Execution speed}

Figure~\ref{fig:speed} shows the statistics of the execution speed of the baseline algorithms in the four systems with both CPU and GPU, and of the commercial solutions.
Considering the baseline algorithms (A1-6), the fastest algorithms are the age and gender estimation algorithms (A5-6), followed by the detection algorithms (A1-2). The tracking algorithms (A3-4) are the slowest ones. All six baseline algorithms run close to real-time (i.e.~30~fps), indicated in the charts by the black dashed line, when considering Systems 1-2 with GPU.
In general, the chart confirms that GPU inference is the fastest, followed by CPU inference with OpenVINO optimization, and the slowest is non-optimized CPU inference.
For instance, A1 with OpenVINO optimization increases its average execution time by 3.6 times (from 0.10~s per frame, in GPU, to 0.36~s per frame, in CPU) in System 1. In the same system, A4 (without OpenVINO optimization) increases its average execution time by 19 times (from 0.08~s per frame, in GPU, to 1.52~s per frame, in CPU). 

\section{Conclusion}
\label{sec:conclusion}

We proposed an open-source benchmark for the evaluation of Anonymous Video Analytics (AVA) for audience measurement and released to the research community the first fully-annotated dataset that enables the evaluation of AVA algorithms. Using this benchmark, we conducted a set of experiments with eight baseline algorithms and two commercial off-the-shelf solutions for the tasks of localization, counting, age, and gender estimation. All the tasks are evaluated in four systems, with CPU and GPU. Results showed that trackers perform better than detectors in all scenarios, that localization algorithms should improve when objects are far/occluded (Figures~\ref{fig:localization} and~\ref{fig:count}), and that the use of higher input frame rate videos do not ensure a better performance.
Further efforts should be made towards the design of holistic tracking solutions that synergistically consider body and face to account for robustness and attention attributes. The performance of age estimation algorithms is limited and results suggest that the performance in estimating the age for younger, [0,18], and older people, [65+], is degraded. This might be due to an existing bias in the datasets used for training age estimation algorithms. Based on the outcomes of the benchmark, future work could explore the design of improved AVA algorithms for age estimation and cumulative count based on multiple object tracking, as well as for attention to evaluate the audience responsiveness to an advertisement.

\section*{Availability of data and materials}
The dataset, benchmark and related open-source codes are available at \href{http://ava.eecs.qmul.ac.uk}{http://ava.eecs.qmul.ac.uk}.

\section*{Competing interests}
Intel Corporation\footnote{{Intel is committed to respecting human rights and avoiding complicity in human rights abuses. See \href{https://www.intel.com/content/www/us/en/policy/policy-human-rights.html}{Intel’s Global Human Rights Principles}. Intel’s products and software are intended only to be used in applications that do not cause or contribute to a violation of an internationally recognized human right.}} provided funding and hardware to support the design and development of the research presented in this manuscript.

\section*{Funding}
The research presented in this paper was supported by Intel Corporation.

\section*{Authors' contributions}
All authors participated in the design of the analytics, performance measures, experiments, and writing of the manuscript.

\section*{Acknowledgments}
We wish to thank Sangeeta Ghangam {Manepalli} for her feedback and Chau Yi Li for her help in developing the online evaluation tool.

\section*{Abbreviations}
In alphabetical order.
AVA: Anonymous Video Analytics; 
COCO: Common Objects in Context;
COE: Cumulative Opportunity Error;
CPE: Cumulative Person Error;
CPU: Central Processing Unit;
FN: False Negative;
FP: False Positive;
GPU: Graphics Processing Unit;
IOU: Intersection Over Union;
MOE: Mean Opportunity Error;
MPE: Mean People Error;
OTS: Opportunity to See;
TCOE: Temporal Cumulative Opportunity Error;
TP: True Positive.

\bibliographystyle{bmc-mathphys} 
\bibliography{main}

\end{document}

%% file: commercial.tex
\begin{figure}[!t]
    \centering
    \setlength{\tabcolsep}{1pt}
    \begin{tabular}{ccc}
    \specialrule{1.2pt}{0.2pt}{1pt}
    \multicolumn{3}{c}{\textbf{Localization}} \\
    \begin{tikzpicture}
        \begin{axis}[
            axis x line=bottom,
            width=0.35\columnwidth,
            height=4cm,
            xmin=0, xmax=7,
            ymajorgrids=true,
            ylabel={Precision},
            boxplot/draw direction=y,
            xtick={1,2,3,4,5,6},
            xticklabels={,,},
            ymin=0,ymax=1,
            tick label style={font=\scriptsize},
            label style={font=\scriptsize},
            ticklabel style={font=\scriptsize},
            ylabel near ticks,
            xticklabel style={rotate=90},
        ]
        \addplot+[boxplot, A1, fill, solid, fill opacity=0.5, boxplot/draw position=1,mark=*, mark options={white,scale=0.5},boxplot/box extend=0.5] table[y=P]{\commretinaDet};
        \addplot+[boxplot, A2, fill, solid, fill opacity=0.5, boxplot/draw position=2,mark=*, mark options={white,scale=0.5},boxplot/box extend=0.5] table[y=P]{\commmtcnnDet};
        \addplot+[boxplot, A3, fill, solid, fill opacity=0.5, boxplot/draw position=3,mark=*, mark options={white,scale=0.5},boxplot/box extend=0.5] table[y=P]{\commdeepsortDet};
        \addplot+[boxplot, A4, fill, solid, fill opacity=0.5, boxplot/draw position=4,mark=*, mark options={white,scale=0.5},boxplot/box extend=0.5] table[y=P]{\commtrmotDet};
        \addplot+[boxplot, C1, fill, solid, fill opacity=0.5, boxplot/draw position=5,mark=*, mark options={white,scale=0.5},boxplot/box extend=0.5] table[y=P]{\commquividiDet};
        \addplot+[boxplot, C2, fill, solid, fill opacity=0.5, boxplot/draw position=6,mark=*, mark options={white,scale=0.5},boxplot/box extend=0.5] table[y=P]{\commmeldcxDet};
        \end{axis}
        \end{tikzpicture}
        &
        \begin{tikzpicture}
        \begin{axis}[
            axis x line=bottom,
            width=0.35\columnwidth,
            height=4cm,
            xmin=0, xmax=7,
            ymajorgrids=true,
            ylabel={Recall},
            boxplot/draw direction=y,
            xtick={1,2,3,4,5,6},
            xticklabels={,,},
            ymin=0,ymax=1,
            yticklabels={,,},
            tick label style={font=\scriptsize},
            label style={font=\scriptsize},
            ticklabel style={font=\scriptsize},
            ylabel near ticks,
            xticklabel style={rotate=90},
        ]
        \addplot+[boxplot, A1, fill, solid, fill opacity=0.5, boxplot/draw position=1,mark=*, mark options={white,scale=0.5},boxplot/box extend=0.5] table[y=R]{\commretinaDet};
        \addplot+[boxplot, A2, fill, solid, fill opacity=0.5, boxplot/draw position=2,mark=*, mark options={white,scale=0.5},boxplot/box extend=0.5] table[y=R]{\commmtcnnDet};
        \addplot+[boxplot, A3, fill, solid, fill opacity=0.5, boxplot/draw position=3,mark=*, mark options={white,scale=0.5},boxplot/box extend=0.5] table[y=R]{\commdeepsortDet};
        \addplot+[boxplot, A4, fill, solid, fill opacity=0.5, boxplot/draw position=4,mark=*, mark options={white,scale=0.5},boxplot/box extend=0.5] table[y=R]{\commtrmotDet};
        \addplot+[boxplot, C1, fill, solid, fill opacity=0.5, boxplot/draw position=5,mark=*, mark options={white,scale=0.5},boxplot/box extend=0.5] table[y=R]{\commquividiDet};
        \addplot+[boxplot, C2, fill, solid, fill opacity=0.5, boxplot/draw position=6,mark=*, mark options={white,scale=0.5},boxplot/box extend=0.5] table[y=R]{\commmeldcxDet};
        \end{axis}
        \end{tikzpicture}
        &
        \begin{tikzpicture}
        \begin{axis}[
            axis x line=bottom,
            width=0.35\columnwidth,
            height=4cm,
            xmin=0, xmax=7,
            ymajorgrids=true,
            ylabel={F1-Score},
            boxplot/draw direction=y,
            xtick={1,2,3,4,5,6},
            xticklabels={,,},
            ymin=0,ymax=1,
            yticklabels={,,},
            tick label style={font=\scriptsize},
            label style={font=\scriptsize},
            ticklabel style={font=\scriptsize},
            ylabel near ticks,
            xticklabel style={rotate=90},
        ]
        \addplot+[boxplot, A1, fill, solid, fill opacity=0.5, boxplot/draw position=1,mark=*, mark options={white,scale=0.5},boxplot/box extend=0.5] table[y=F]{\commretinaDet};
        \addplot+[boxplot, A2, fill, solid, fill opacity=0.5, boxplot/draw position=2,mark=*, mark options={white,scale=0.5},boxplot/box extend=0.5] table[y=F]{\commmtcnnDet};
        \addplot+[boxplot, A3, fill, solid, fill opacity=0.5, boxplot/draw position=3,mark=*, mark options={white,scale=0.5},boxplot/box extend=0.5] table[y=F]{\commdeepsortDet};
        \addplot+[boxplot, A4, fill, solid, fill opacity=0.5, boxplot/draw position=4,mark=*, mark options={white,scale=0.5},boxplot/box extend=0.5] table[y=F]{\commtrmotDet};
        \addplot+[boxplot, C1, fill, solid, fill opacity=0.5, boxplot/draw position=5,mark=*, mark options={white,scale=0.5},boxplot/box extend=0.5] table[y=F]{\commquividiDet};
        \addplot+[boxplot, C2, fill, solid, fill opacity=0.5, boxplot/draw position=6,mark=*, mark options={white,scale=0.5},boxplot/box extend=0.5] table[y=F]{\commmeldcxDet};
        \end{axis}
        \end{tikzpicture}
        \\
        \specialrule{1.2pt}{0.2pt}{1pt}
        \multicolumn{3}{c}{\textbf{Count}}
        \\
        \begin{tikzpicture}
        \begin{axis}[
            axis x line=bottom,
            width=0.35\columnwidth,
            height=4cm,
            xmin=0, xmax=7,
            ylabel={MOE},
            ymajorgrids=true,
            boxplot/draw direction=y,
            xtick={1,2,3,4,5,6},
            xticklabels={,,},
            ymin=0,ymax=10,
            tick label style={font=\scriptsize},
            label style={font=\scriptsize},
            ticklabel style={font=\scriptsize},
            ylabel near ticks,
            xticklabel style={rotate=90},
        ]
        \addplot+[boxplot, A1, fill, solid, fill opacity=0.5, boxplot/draw position=1,mark=*, mark options={white,scale=0.5},boxplot/box extend=0.5] table[y=MOE]{\commretina};
        \addplot+[boxplot, A2, fill, solid, fill opacity=0.5, boxplot/draw position=2,mark=*, mark options={white,scale=0.5},boxplot/box extend=0.5] table[y=MOE]{\commmtcnn};
        \addplot+[boxplot, A3, fill, solid, fill opacity=0.5, boxplot/draw position=3,mark=*, mark options={white,scale=0.5},boxplot/box extend=0.5] table[y=MOE]{\commdeepsort};
        \addplot+[boxplot, A4, fill, solid, fill opacity=0.5, boxplot/draw position=4,mark=*, mark options={white,scale=0.5},boxplot/box extend=0.5] table[y=MOE]{\commtrmot};
        \addplot+[boxplot, C1, fill, solid, fill opacity=0.5, boxplot/draw position=5,mark=*, mark options={white,scale=0.5},boxplot/box extend=0.5] table[y=MOE]{\commquividi};
        \addplot+[boxplot, C2, fill, solid, fill opacity=0.5, boxplot/draw position=6,mark=*, mark options={white,scale=0.5},boxplot/box extend=0.5] table[y=MOE]{\commmeldcx};
        \end{axis}
        \end{tikzpicture}
        &
        \begin{tikzpicture}
            \begin{axis}[
                axis x line=bottom,
                width=0.35\columnwidth,
                height=4cm,
                xmin=0, xmax=7,
                ylabel={MPE},
                ymajorgrids=true,
                boxplot/draw direction=y,
                xtick={1,2,3,4,5,6},
                xticklabels={,,},
                ymin=0,ymax=10,
                tick label style={font=\scriptsize},
                label style={font=\scriptsize},
                ticklabel style={font=\scriptsize},
                ylabel near ticks,
                xticklabel style={rotate=90},
            ]
            \addplot+[boxplot, A1, fill, solid, fill opacity=0.5, boxplot/draw position=1,mark=*, mark options={white,scale=0.5},boxplot/box extend=0.5] table[y=MRE]{\commretina};
            \addplot+[boxplot, A2, fill, solid, fill opacity=0.5, boxplot/draw position=2,mark=*, mark options={white,scale=0.5},boxplot/box extend=0.5] table[y=MRE]{\commmtcnn};
            \addplot+[boxplot, A3, fill, solid, fill opacity=0.5, boxplot/draw position=3,mark=*, mark options={white,scale=0.5},boxplot/box extend=0.5] table[y=MRE]{\commdeepsort};
            \addplot+[boxplot, A4, fill, solid, fill opacity=0.5, boxplot/draw position=4,mark=*, mark options={white,scale=0.5},boxplot/box extend=0.5] table[y=MRE]{\commtrmot};
            \addplot+[boxplot, C1, fill, solid, fill opacity=0.5, boxplot/draw position=5,mark=*, mark options={white,scale=0.5},boxplot/box extend=0.5] table[y=MRE]{\commquividi};
            \addplot+[boxplot, C2, fill, solid, fill opacity=0.5, boxplot/draw position=6,mark=*, mark options={white,scale=0.5},boxplot/box extend=0.5] table[y=MRE]{\commmeldcx};
            \end{axis}
            \end{tikzpicture}
        &
        \begin{tikzpicture}
            \begin{axis}[
                axis x line=bottom,
                width=0.35\columnwidth,
                height=4cm,
                xmin=0, xmax=8,
                ylabel={COE},
                ymajorgrids=true,
                boxplot/draw direction=y,
                xtick={1,2,3,4,5,6,7},
                xticklabels={,,},
                ymin=0,ymax=15,
                tick label style={font=\scriptsize},
                label style={font=\scriptsize},
                ticklabel style={font=\scriptsize},
                ylabel near ticks,
                xticklabel style={rotate=90},
            ]
            \addplot+[boxplot, A1, fill, solid, fill opacity=0.5, boxplot/draw position=1,mark=*, mark options={white,scale=0.5},boxplot/box extend=0.5] table[y=COE]{\commretina};
            \addplot+[boxplot, A2, fill, solid, fill opacity=0.5, boxplot/draw position=2,mark=*, mark options={white,scale=0.5},boxplot/box extend=0.5] table[y=COE]{\commmtcnn};
            \addplot+[boxplot, A3, fill, solid, fill opacity=0.5, boxplot/draw position=3,mark=*, mark options={white,scale=0.5},boxplot/box extend=0.5] table[y=COE]{\commdeepsort};
            \addplot+[boxplot, A4, fill, solid, fill opacity=0.5, boxplot/draw position=4,mark=*, mark options={white,scale=0.5},boxplot/box extend=0.5] table[y=COE]{\commtrmot};
            \addplot+[boxplot, C1, fill, solid, fill opacity=0.5, boxplot/draw position=5,mark=*, mark options={white,scale=0.5},boxplot/box extend=0.5] table[y=COE]{\commquividi};
            \addplot+[boxplot, C2, fill, solid, fill opacity=0.5, boxplot/draw position=6,mark=*, mark options={white,scale=0.5},boxplot/box extend=0.5] table[y=COE]{\commmeldcx};
            \addplot+[boxplot, C2, fill, solid, fill opacity=0.5, boxplot/draw position=7,mark=*, mark options={white,scale=0.5},boxplot/box extend=0.5] table[y=CPE]{\commmeldcx}; 
            \end{axis}
        \end{tikzpicture} 
        \\
        \specialrule{1.2pt}{0.2pt}{1pt}
        \multicolumn{3}{c}{\textbf{Age}}
        \\
        \multicolumn{3}{c}{
        \begin{tikzpicture}
        \begin{axis}[
            axis x line=bottom,
            width=0.85\columnwidth,
            height=4cm,
            xmin=0, xmax=20,
            ymajorgrids=true,
            ylabel={F1-Score},
            boxplot/draw direction=y,
            xtick={5,10,15},
            xticklabels={,,},
            ymin=0,ymax=1,
            tick label style={font=\scriptsize},
            label style={font=\scriptsize},
            ticklabel style={font=\scriptsize},
            ylabel near ticks,
        ]
        \addplot+[boxplot, A5, fill, solid, fill opacity=0.5, boxplot/draw position=1,mark=*, mark options={white,scale=0.5},boxplot/box extend=0.5] table[y=F0]{\commfacelibAge};
        \addplot+[boxplot, A6, fill, solid, fill opacity=0.5, boxplot/draw position=2,mark=*, mark options={white,scale=0.5},boxplot/box extend=0.5] table[y=F0]{\commdexAge};
        \addplot+[boxplot, C1, fill, solid, fill opacity=0.5, boxplot/draw position=3,mark=*, mark options={white,scale=0.5},boxplot/box extend=0.5] table[y=F0]{\commquividiAge};
        \addplot+[boxplot, C2, fill, solid, fill opacity=0.5, boxplot/draw position=4,mark=*, mark options={white,scale=0.5},boxplot/box extend=0.5] table[y=F0]{\commmeldcxAge};
        \addplot+[boxplot, A5, fill, solid, fill opacity=0.5, boxplot/draw position=6,mark=*, mark options={white,scale=0.5},boxplot/box extend=0.5] table[y=F1]{\commfacelibAge};
        \addplot+[boxplot, A6, fill, solid, fill opacity=0.5, boxplot/draw position=7,mark=*, mark options={white,scale=0.5},boxplot/box extend=0.5] table[y=F1]{\commdexAge};
        \addplot+[boxplot, C1, fill, solid, fill opacity=0.5, boxplot/draw position=8,mark=*, mark options={white,scale=0.5},boxplot/box extend=0.5] table[y=F1]{\commquividiAge};
        \addplot+[boxplot, C2, fill, solid, fill opacity=0.5, boxplot/draw position=9,mark=*, mark options={white,scale=0.5},boxplot/box extend=0.5] table[y=F1]{\commmeldcxAge};
        \addplot+[boxplot, A5, fill, solid, fill opacity=0.5, boxplot/draw position=11,mark=*, mark options={white,scale=0.5},boxplot/box extend=0.5] table[y=F2]{\commfacelibAge};
        \addplot+[boxplot, A6, fill, solid, fill opacity=0.5, boxplot/draw position=12,mark=*, mark options={white,scale=0.5},boxplot/box extend=0.5] table[y=F2]{\commdexAge};
        \addplot+[boxplot, C1, fill, solid, fill opacity=0.5, boxplot/draw position=13,mark=*, mark options={white,scale=0.5},boxplot/box extend=0.5] table[y=F2]{\commquividiAge};
        \addplot+[boxplot, C2, fill, solid, fill opacity=0.5, boxplot/draw position=14,mark=*, mark options={white,scale=0.5},boxplot/box extend=0.5] table[y=F2]{\commmeldcxAge};
        \addplot+[boxplot, A5, fill, solid, fill opacity=0.5, boxplot/draw position=16,mark=*, mark options={white,scale=0.5},boxplot/box extend=0.5] table[y=F3]{\commfacelibAge};
        \addplot+[boxplot, A6, fill, solid, fill opacity=0.5, boxplot/draw position=17,mark=*, mark options={white,scale=0.5},boxplot/box extend=0.5] table[y=F3]{\commdexAge};
        \addplot+[boxplot, C1, fill, solid, fill opacity=0.5, boxplot/draw position=18,mark=*, mark options={white,scale=0.5},boxplot/box extend=0.5] table[y=F3]{\commquividiAge};
        \addplot+[boxplot, C2, fill, solid, fill opacity=0.5, boxplot/draw position=19,mark=*, mark options={white,scale=0.5},boxplot/box extend=0.5] table[y=F3]{\commmeldcxAge};
        \end{axis}
        \begin{axis}[
            axis x line*=top,
            width=0.85\columnwidth,
            height=4cm,
            xmin=0, xmax=20,
            boxplot/draw direction=y,
            xtick={2.5,7.5,12.5,17.5},
            xticklabels={{[0,18]}, {[19,34]}, {[35,65]}, {[65+]}},
            yticklabels={,,},
            ymin=0,ymax=1,
            tick label style={font=\scriptsize},
            label style={font=\scriptsize},
            ticklabel style={font=\scriptsize},
            ylabel near ticks
        ]
        \end{axis}
        \end{tikzpicture}
        }
        \\
        \specialrule{1.2pt}{0.2pt}{1pt}
        \multicolumn{3}{c}{\textbf{Gender}}
        \\
        \multicolumn{3}{c}{
        \begin{tikzpicture}
        \begin{axis}[
            axis x line=bottom,
            width=0.85\columnwidth,
            height=4cm,
            xmin=0, xmax=10,
            ymajorgrids=true,
            ylabel={F1-Score},
            boxplot/draw direction=y,
            xtick={5},
            xticklabels={,,},
            ymin=0,ymax=1,
            tick label style={font=\scriptsize},
            label style={font=\scriptsize},
            ticklabel style={font=\scriptsize},
            ylabel near ticks,
        ]
        \addplot+[boxplot, A5, fill, solid, fill opacity=0.5, boxplot/draw position=1,mark=*, mark options={white,scale=0.5},boxplot/box extend=0.5] table[y=Fmale]{\commfacelibGen};
        \addplot+[boxplot, A6, fill, solid, fill opacity=0.5, boxplot/draw position=2,mark=*, mark options={white,scale=0.5},boxplot/box extend=0.5] table[y=Fmale]{\commdexGen};
        \addplot+[boxplot, C1, fill, solid, fill opacity=0.5, boxplot/draw position=3,mark=*, mark options={white,scale=0.5},boxplot/box extend=0.5] table[y=Fmale]{\commquividiGen};
        \addplot+[boxplot, C2, fill, solid, fill opacity=0.5, boxplot/draw position=4,mark=*, mark options={white,scale=0.5},boxplot/box extend=0.5] table[y=Fmale]{\commmeldcxGen};
        \addplot+[boxplot, A5, fill, solid, fill opacity=0.5, boxplot/draw position=6,mark=*, mark options={white,scale=0.5},boxplot/box extend=0.5] table[y=Ffemale]{\commfacelibGen};
        \addplot+[boxplot, A6, fill, solid, fill opacity=0.5, boxplot/draw position=7,mark=*, mark options={white,scale=0.5},boxplot/box extend=0.5] table[y=Ffemale]{\commdexGen};
        \addplot+[boxplot, C1, fill, solid, fill opacity=0.5, boxplot/draw position=8,mark=*, mark options={white,scale=0.5},boxplot/box extend=0.5] table[y=Ffemale]{\commquividiGen};
        \addplot+[boxplot, C2, fill, solid, fill opacity=0.5, boxplot/draw position=9,mark=*, mark options={white,scale=0.5},boxplot/box extend=0.5] table[y=Ffemale]{\commmeldcxGen};
        \end{axis}
        \begin{axis}[
            axis x line*=top,
            width=0.85\columnwidth,
            height=4cm,
            xmin=0, xmax=10,
            boxplot/draw direction=y,
            xtick={2.5,7.5},
            xticklabels={\textit{Male},\textit{Female}},
            yticklabels={,,},
            ymin=0,ymax=1,
            tick label style={font=\scriptsize},
            label style={font=\scriptsize},
            ticklabel style={font=\scriptsize},
            ylabel near ticks
        ]
        \end{axis}
        \end{tikzpicture}
        }
    \end{tabular}
    \caption{Anonymous Video Analytics comparison of
    Algorithm~1~(A1:~\protect\tikz \protect\draw[fill=A1,draw=none] (0,0) rectangle (1.ex,1.ex);),
    Algorithm~2~(A2:~\protect\tikz \protect\draw[fill=A2,draw=none] (0,0) rectangle (1.ex,1.ex);),
    Algorithm~3~(A3:~\protect\tikz \protect\draw[fill=A3,draw=none] (0,0) rectangle (1.ex,1.ex);),
    Algorithm~4~(A4:~\protect\tikz \protect\draw[fill=A4,draw=none] (0,0) rectangle (1.ex,1.ex););
    Algorithm~5~(A5:~\protect\tikz \protect\draw[fill=A5,draw=none] (0,0) rectangle (1.ex,1.ex);),
    Algorithm~6~(A6:~\protect\tikz \protect\draw[fill=A6,draw=none] (0,0) rectangle (1.ex,1.ex);); and commercial solutions
    Commercial off-the-shelf~1 (C1:~\protect\tikz \protect\draw[fill=C1,draw=none] (0,0) rectangle (1.ex,1.ex);), and
    Commercial off-the-shelf~2 (C2:~\protect\tikz \protect\draw[fill=C2,draw=none] (0,0) rectangle (1.ex,1.ex);), accounting for localization in terms of precision, recall, and F1-Score; people counting in terms of Mean Opportunity Error (MOE), Mean People Error (MPE), and Cumulative Opportunity Error (COE); and age and gender estimation in terms of F1-Score.
    All algorithms have been executed with the same video inputs (at 30~frames per second).
    A1-6 are run in System 1 with GPU, and C1-2 are run in CPU.
    Note that A1-2 obtain a COE several orders of magnitude larger than the rest of the algorithms; thus, these results are out of the chart ranges for easing the visualization.
    The company developing C2 defines OTS as any person within the field of view, differently than the definition proposed in this paper. To have this into consideration, we show two bar plots for C2. The one with the highest error corresponds to using the definition proposed in this paper, and the other one using the definition proposed by C2 (i.e.~CPE, Eq.~\ref{eq:cpe}, is reported).}
    \label{fig:results_commercial}
\end{figure}

\begin{figure}[!t]
    \centering
    \setlength{\tabcolsep}{-2pt}
    \begin{tabular}{cccc}
    \begin{tikzpicture}
        \begin{axis}[
            xmode=log,
            axis x line=bottom,
            width=0.35\columnwidth,
            height=0.4\columnwidth,
            xmin=10, xmax=120,
            ymin=0, ymax=60,
            ylabel={TCOE},
            ymajorgrids=true,
            xtick={10,30,60,120},
            xticklabels={10,30,60,120},
            xlabel={Segm. [secs]},
            tick label style={font=\scriptsize},
            label style={font=\scriptsize},
            ticklabel style={font=\scriptsize},
            ylabel near ticks,
            xticklabel style={rotate=90},
        ]
        \addplot [name path=upper,draw=none] table[x=RANGE,y expr=\thisrow{TCOEmean}+\thisrow{TCOEstd}] {\commtcoedeepsort};
        \addplot [name path=lower,draw=none] table[x=RANGE,y expr=\thisrow{TCOEmean}-\thisrow{TCOEstd}] {\commtcoedeepsort};
        \addplot [fill=A3, opacity=0.1] fill between[of=upper and lower];
        \addplot+[solid, color=A3, mark=none, style={thick}] table[x=RANGE, y=TCOEmean]{\commtcoedeepsort};
        \end{axis}
    \end{tikzpicture}
    &
    \begin{tikzpicture}
        \begin{axis}[
            xmode=log,
            axis x line=bottom,
            width=0.35\columnwidth,
            height=0.4\columnwidth,
            xmin=10, xmax=120,
            ymin=0, ymax=60,
            ymajorgrids=true,
            xtick={10,30,60,120},
            xticklabels={10,30,60,120},
            yticklabels={,,},
            xlabel={Segm. [secs]},
            tick label style={font=\scriptsize},
            label style={font=\scriptsize},
            ticklabel style={font=\scriptsize},
            ylabel near ticks,
            xticklabel style={rotate=90},
        ]
        \addplot [name path=upper,draw=none] table[x=RANGE,y expr=\thisrow{TCOEmean}+\thisrow{TCOEstd}] {\commtcoetrmot};
        \addplot [name path=lower,draw=none] table[x=RANGE,y expr=\thisrow{TCOEmean}-\thisrow{TCOEstd}] {\commtcoetrmot};
        \addplot [fill=A4, opacity=0.1] fill between[of=upper and lower];
        \addplot+[solid, color=A4, mark=none, style={thick}] table[x=RANGE, y=TCOEmean]{\commtcoetrmot};
        \end{axis}
    \end{tikzpicture}
    &
    \begin{tikzpicture}
        \begin{axis}[
            xmode=log,
            axis x line=bottom,
            width=0.35\columnwidth,
            height=0.4\columnwidth,
            xmin=10, xmax=120,
            ymin=0, ymax=60,
            ymajorgrids=true,
            xtick={10,30,60,120},
            xticklabels={10,30,60,120},
            yticklabels={,,},
            xlabel={Segm. [secs]},
            tick label style={font=\scriptsize},
            label style={font=\scriptsize},
            ticklabel style={font=\scriptsize},
            ylabel near ticks,
            xticklabel style={rotate=90},
        ]
        \addplot [name path=upper,draw=none] table[x=RANGE,y expr=\thisrow{TCOEmean}+\thisrow{TCOEstd}] {\commtcoequividi};
        \addplot [name path=lower,draw=none] table[x=RANGE,y expr=\thisrow{TCOEmean}-\thisrow{TCOEstd}] {\commtcoequividi};
        \addplot [fill=C1, opacity=0.1] fill between[of=upper and lower];
        \addplot+[solid, color=C1, mark=none, style={thick}] table[x=RANGE, y=TCOEmean]{\commtcoequividi};
        \end{axis}
    \end{tikzpicture}
    &
    \begin{tikzpicture}
        \begin{axis}[
            xmode=log,
            axis x line=bottom,
            width=0.35\columnwidth,
            height=0.4\columnwidth,
            xmin=10, xmax=120,
            ymin=0, ymax=60,
            ymajorgrids=true,
            xtick={10,30,60,120},
            xticklabels={10,30,60,120},
            yticklabels={,,},
            xlabel={Segm. [secs]},
            tick label style={font=\scriptsize},
            label style={font=\scriptsize},
            ticklabel style={font=\scriptsize},
            ylabel near ticks,
            xticklabel style={rotate=90},
        ]
        \addplot [name path=upper,draw=none] table[x=RANGE,y expr=\thisrow{TCOEmean}+\thisrow{TCOEstd}] {\commmtcoemeldcx};
        \addplot [name path=lower,draw=none] table[x=RANGE,y expr=\thisrow{TCOEmean}-\thisrow{TCOEstd}] {\commmtcoemeldcx};
        \addplot [fill=C2, opacity=0.1] fill between[of=upper and lower];
        \addplot+[solid, color=C2, mark=none, style={thick}] table[x=RANGE, y=TCOEmean]{\commmtcoemeldcx};
        \addplot [name path=upper,draw=none] table[x=RANGE,y expr=\thisrow{TCPEmean}+\thisrow{TCPEstd}] {\commmtcoemeldcx};
        \addplot [name path=lower,draw=none] table[x=RANGE,y expr=\thisrow{TCPEmean}-\thisrow{TCPEstd}] {\commmtcoemeldcx};
        \addplot [fill=C2, opacity=0.1] fill between[of=upper and lower];
        \addplot+[solid, color=C2, mark=none, style={thick}] table[x=RANGE, y=TCPEmean]{\commmtcoemeldcx};
        \end{axis}
    \end{tikzpicture}
    \end{tabular}
    \caption{People counting results in terms of Temporal Cumulative Opportunity Error (TCOE) for different segment duration (segm.) with 
    Algorithm~3~(A3:~{\protect\raisebox{2pt}{\protect\tikz \protect\draw[A3,line width=1.5] (0,0) -- (0.3,0);}}),
    Algorithm~4~(A4:~{\protect\raisebox{2pt}{\protect\tikz \protect\draw[A4,line width=1.5] (0,0) -- (0.3,0);}}); and
    Commercial off-the-shelf~1~(C1:~{\protect\raisebox{2pt}{\protect\tikz \protect\draw[C1,line width=1.5] (0,0) -- (0.3,0);}}), and
    Commercial off-the-shelf~2~(C2:~{\protect\raisebox{2pt}{\protect\tikz \protect\draw[C2,line width=1.5] (0,0) -- (0.3,0);}}).
    The x axes are in logarithmic scale. We do not show the results for detectors (A1-2) as they obtain a TCOE several orders of magnitude larger than the rest of the algorithms.
    The company developing C2 defines OTS as any person within the field of view, differently than the definition proposed in this paper. To have this into consideration, we show two line plots for C2. The one with the highest error corresponds to using the definition proposed in this paper, and the other line using the definition proposed by C2.}
    \label{fig:count_commercial_TCOE}
\end{figure}

%% file: speed.tex
\begin{figure*}[!t]
    \centering
    \setlength{\tabcolsep}{1pt}
    \begin{tabular}{cccccc}
    \begin{tikzpicture}
        \begin{axis}[
            ymode=log,
            axis x line=bottom,
            width=0.25\textwidth,
            height=7cm,
            xmin=0, xmax=18,
            ymin=0.0018,ymax=20,
            title={System 1},
            ymajorgrids=false,
            boxplot/draw direction=y,
            xticklabels={,,},
            ylabel={Seconds per frame},
            tick label style={font=\scriptsize},
            label style={font=\scriptsize},
            ticklabel style={font=\scriptsize},
            ylabel near ticks,
        ]
        \addplot+[boxplot, A1, fill, solid, fill opacity=0.5, boxplot/draw position=1,mark=*, mark options={white,scale=1.0},boxplot/box extend=0.5] table[y=Speed]{\retinagpuSA};
        \addplot+[boxplot, A1, solid, boxplot/draw position=2,mark=*, mark options={white,scale=1.0},boxplot/box extend=0.5] table[y=Speed]{\retinacpuSA};
        \addplot+[boxplot, A2, fill, solid, fill opacity=0.5, boxplot/draw position=4,mark=*, mark options={white,scale=1.0},boxplot/box extend=0.5] table[y=Speed]{\mtcnngpuSA};
        \addplot+[boxplot, A2, solid, boxplot/draw position=5,mark=*, mark options={white,scale=1.0},boxplot/box extend=0.5] table[y=Speed]{\mtcnncpuSA};
        \addplot+[boxplot, A3, fill, solid, fill opacity=0.5, boxplot/draw position=7,mark=*, mark options={white,scale=1.0},boxplot/box extend=0.5] table[y=Speed]{\deepsortgpuSA};
        \addplot+[boxplot, A3, solid, boxplot/draw position=8,mark=*, mark options={white,scale=1.0},boxplot/box extend=0.5] table[y=Speed]{\deepsortcpuSA};
        \addplot+[boxplot, A4, fill, solid, fill opacity=0.5, boxplot/draw position=10,mark=*, mark options={white,scale=1.0},boxplot/box extend=0.5] table[y=Speed]{\trmotgpuSA};
        \addplot+[boxplot, A4, solid, boxplot/draw position=11, mark=*, mark options={white,scale=1.0},boxplot/box extend=0.5] table[y=Speed]{\trmotcpuSA};
        \addplot+[boxplot, A5, fill, solid, fill opacity=0.5, boxplot/draw position=13,mark=*, mark options={white,scale=1.0},boxplot/box extend=0.5] table[y=Speed]{\facelibgpuAgeSA};
        \addplot+[boxplot, A5, solid, fill opacity=0.5, boxplot/draw position=14,mark=*, mark options={white,scale=1.0},boxplot/box extend=0.5] table[y=Speed]{\facelibcpuAgeSA};
        \addplot+[boxplot, A6, fill, solid, fill opacity=0.5, boxplot/draw position=16,mark=*, mark options={white,scale=1.0},boxplot/box extend=0.5] table[y=Speed]{\dexgpuAgeSA};
        \addplot+[boxplot, A6, solid, fill opacity=0.5, boxplot/draw position=17,mark=*, mark options={white,scale=1.0},boxplot/box extend=0.5] table[y=Speed]{\dexcpuAgeSA};
        \addplot[domain=0:18, thick, dashed, black]{0.03333};
        \end{axis}
        \end{tikzpicture}
        &
        \begin{tikzpicture}
        \begin{axis}[
            ymode=log,
            axis x line=bottom,
            width=0.25\textwidth,
            height=7cm,
            xmin=0, xmax=18,
            ymin=0.0018,ymax=20,
            title={System 2},
            ymajorgrids=false,
            boxplot/draw direction=y,
            xticklabels={,,},
            yticklabels={,,},
            tick label style={font=\scriptsize},
            label style={font=\scriptsize},
            ticklabel style={font=\scriptsize},
            ylabel near ticks,
        ]
        \addplot+[boxplot, A1, fill, solid, fill opacity=0.5, boxplot/draw position=1,mark=*, mark options={white,scale=1.0},boxplot/box extend=0.5] table[y=Speed]{\retinagpuSB};
        \addplot+[boxplot, A1, solid, boxplot/draw position=2,mark=*, mark options={white,scale=1.0},boxplot/box extend=0.5] table[y=Speed]{\retinacpuSB};
        \addplot+[boxplot, A2, fill, solid, fill opacity=0.5, boxplot/draw position=4,mark=*, mark options={white,scale=1.0},boxplot/box extend=0.5] table[y=Speed]{\mtcnngpuSB};
        \addplot+[boxplot, A2, solid, boxplot/draw position=5,mark=*, mark options={white,scale=1.0},boxplot/box extend=0.5] table[y=Speed]{\mtcnncpuSB};
        \addplot+[boxplot, A3, fill, solid, fill opacity=0.5, boxplot/draw position=7,mark=*, mark options={white,scale=1.0},boxplot/box extend=0.5] table[y=Speed]{\deepsortgpuSB};
        \addplot+[boxplot, A3, solid, boxplot/draw position=8,mark=*, mark options={white,scale=1.0},boxplot/box extend=0.5] table[y=Speed]{\deepsortcpuSB};
        \addplot+[boxplot, A4, fill, solid, fill opacity=0.5, boxplot/draw position=10,mark=*, mark options={white,scale=1.0},boxplot/box extend=0.5] table[y=Speed]{\trmotgpuSB};
        \addplot+[boxplot, A4, solid, boxplot/draw position=11, mark=*, mark options={white,scale=1.0},boxplot/box extend=0.5] table[y=Speed]{\trmotcpuSB};
        \addplot+[boxplot, A5, fill, solid, fill opacity=0.5, boxplot/draw position=13,mark=*, mark options={white,scale=1.0},boxplot/box extend=0.5] table[y=Speed]{\facelibgpuAgeSB};
        \addplot+[boxplot, A5, solid, fill opacity=0.5, boxplot/draw position=14,mark=*, mark options={white,scale=1.0},boxplot/box extend=0.5] table[y=Speed]{\facelibcpuAgeSB};
        \addplot+[boxplot, A6, fill, solid, fill opacity=0.5, boxplot/draw position=16,mark=*, mark options={white,scale=1.0},boxplot/box extend=0.5] table[y=Speed]{\dexgpuAgeSB};
        \addplot+[boxplot, A6, solid, fill opacity=0.5, boxplot/draw position=17,mark=*, mark options={white,scale=1.0},boxplot/box extend=0.5] table[y=Speed]{\dexcpuAgeSB};
        \addplot[domain=0:18, thick, dashed, black]{0.03333};
        \end{axis}
        \end{tikzpicture}
        &
        \begin{tikzpicture}
        \begin{axis}[
            ymode=log,
            axis x line=bottom,
            width=0.25\textwidth,
            height=7cm,
            xmin=0, xmax=18,
            ymin=0.0018,ymax=20,
            title={System 3},
            ymajorgrids=false,
            boxplot/draw direction=y,
            xticklabels={,,},
            yticklabels={,,},
            tick label style={font=\scriptsize},
            label style={font=\scriptsize},
            ticklabel style={font=\scriptsize},
            ylabel near ticks,
        ]
        \addplot+[boxplot, A1, solid, boxplot/draw position=2,mark=*, mark options={white,scale=1.0},boxplot/box extend=0.5] table[y=Speed]{\retinacpuSC};
        \addplot+[boxplot, A2, solid, boxplot/draw position=5,mark=*, mark options={white,scale=1.0},boxplot/box extend=0.5] table[y=Speed]{\mtcnncpuSC};
        \addplot+[boxplot, A3, solid, boxplot/draw position=8,mark=*, mark options={white,scale=1.0},boxplot/box extend=0.5] table[y=Speed]{\deepsortcpuSC};
        \addplot+[boxplot, A4, solid, boxplot/draw position=11, mark=*, mark options={white,scale=1.0},boxplot/box extend=0.5] table[y=Speed]{\trmotcpuSC};
        %
        \addplot+[boxplot, A5, solid, fill opacity=0.5, boxplot/draw position=14,mark=*, mark options={white,scale=1.0},boxplot/box extend=0.5] table[y=Speed]{\facelibcpuAgeSC};
        \addplot+[boxplot, A6, solid, fill opacity=0.5, boxplot/draw position=17,mark=*, mark options={white,scale=1.0},boxplot/box extend=0.5] table[y=Speed]{\dexcpuAgeSC};
        \addplot[domain=0:18, thick, dashed, black]{0.03333};
        \end{axis}
        \end{tikzpicture}
        &
        \begin{tikzpicture}
        \begin{axis}[
            ymode=log,
            axis x line=bottom,
            width=0.25\textwidth,
            height=7cm,
            xmin=0, xmax=18,
            ymin=0.0018,ymax=20,
            title={System 4},
            ymajorgrids=false,
            boxplot/draw direction=y,
            xticklabels={,,},
            yticklabels={,,},
            tick label style={font=\scriptsize},
            label style={font=\scriptsize},
            ticklabel style={font=\scriptsize},
            ylabel near ticks,
        ]
        \addplot+[boxplot, A1, fill, solid, fill opacity=0.5, boxplot/draw position=1,mark=*, mark options={white,scale=1.0},boxplot/box extend=0.5] table[y=Speed]{\retinagpuSD};
        \addplot+[boxplot, A2, fill, solid, fill opacity=0.5, boxplot/draw position=4,mark=*, mark options={white,scale=1.0},boxplot/box extend=0.5] table[y=Speed]{\mtcnngpuSD};
        \addplot+[boxplot, A3, fill, solid, fill opacity=0.5, boxplot/draw position=7,mark=*, mark options={white,scale=1.0},boxplot/box extend=0.5] table[y=Speed]{\deepsortgpuSD};
        \addplot+[boxplot, A4, fill, solid, fill opacity=0.5, boxplot/draw position=10,mark=*, mark options={white,scale=1.0},boxplot/box extend=0.5] table[y=Speed]{\trmotgpuSD};
        %
        \addplot+[boxplot, A5, fill, solid, fill opacity=0.5, boxplot/draw position=13,mark=*, mark options={white,scale=1.0},boxplot/box extend=0.5] table[y=Speed]{\facelibgpuAgeSD};
        \addplot[domain=0:18, thick, dashed, black]{0.03333};
        \end{axis}
        \end{tikzpicture}
        &
        \begin{tikzpicture}
        \begin{axis}[
            ymode=log,
            axis x line=bottom,
            width=0.25\textwidth,
            height=7cm,
            xmin=0, xmax=3,
            ymin=0.0018,ymax=20,
            title={C1-2 system},
            ymajorgrids=false,
            boxplot/draw direction=y,
            xticklabels={,,},
            yticklabels={,,},
            tick label style={font=\scriptsize},
            label style={font=\scriptsize},
            ticklabel style={font=\scriptsize},
            ylabel near ticks,
        ]
        \addplot+[boxplot, C1, fill, solid, fill opacity=0.5, boxplot/draw position=1,mark=*, mark options={white,scale=1.0},boxplot/box extend=0.5] table[y=Speed]{\commquividi};
        \addplot+[boxplot, C2, fill, solid, fill opacity=0.5, boxplot/draw position=2,mark=*, mark options={white,scale=1.0},boxplot/box extend=0.5] table[y=Speed]{\commmeldcx};
        \addplot[domain=0:3, thick, dashed, black]{0.03333};
        \end{axis}
        \end{tikzpicture}
    \end{tabular}
    \caption{Per-frame execution speed of baseline algorithms:
    Algorithm~1 (A1:~\protect\tikz \protect\draw[fill=A1,draw=none] (0,0) rectangle (1.ex,1.ex);),
    Algorithm~2 (A2:~\protect\tikz \protect\draw[fill=A2,draw=none] (0,0) rectangle (1.ex,1.ex);),
    Algorithm~3 (A3:~\protect\tikz \protect\draw[fill=A3,draw=none] (0,0) rectangle (1.ex,1.ex);), 
    Algorithm~4 (A4:~\protect\tikz \protect\draw[fill=A4,draw=none] (0,0) rectangle (1.ex,1.ex);),
    Algorithm~5 (A5:~\protect\tikz \protect\draw[fill=A5,draw=none] (0,0) rectangle (1.ex,1.ex);),
    Algorithm~6 (A6:~\protect\tikz \protect\draw[fill=A6,draw=none] (0,0) rectangle (1.ex,1.ex););
    and
    Commercial off-the-shelf~1 (C1:~\protect\tikz \protect\draw[fill=C1,draw=none] (0,0) rectangle (1.ex,1.ex);), and
    Commercial off-the-shelf~2 (C2:~\protect\tikz \protect\draw[fill=C2,draw=none] (0,0) rectangle (1.ex,1.ex);); across all videos of the AVA dataset.
    When compatible with the system, the results are reported with inference in GPU (filled boxes) and CPU (non-filled boxes). The horizontal dashed line indicates~30~fps.
    Commercial solutions are run on a subset of 6 videos and in external systems equipped with an Intel i7 CPU.
    \magenta{}
    }
    \label{fig:speed}
\end{figure*}

%% file: pervideo.tex
\begin{figure}[!t]
    \centering
    \setlength{\tabcolsep}{0pt}
    \begin{tabular}{cc}
    \specialrule{1.2pt}{0.2pt}{1pt}
    \multicolumn{2}{c}{\textbf{Count}} \\
    \cmidrule(lr){1-2}
    {Instantaneous} & {Cumulative} \\
    \specialrule{1.2pt}{0.2pt}{1pt}
    \begin{tikzpicture}
    \begin{axis}[
        width=0.53\columnwidth,
        height=3.5cm,
        ymin=0,ymax=20,
        xmin=0,xmax=10000,
        ylabel={{Airport-1}},
        label style={font=\footnotesize},
        tick label style={font=\footnotesize},
        ylabel near ticks,
        xlabel near ticks,
        xticklabels={,,},
        xtick scale label code/.code={},
    ]
        \addplot+[solid, intel, mark=none, style={thick}] table[x=Frame, y=actInst]{\SAsequenceA};
        \addplot+[solid, A1, mark=none, fill opacity=0.3] table[x=Frame, y=estInst0]{\SAsequenceA};
        \addplot+[solid, A2, mark=none, fill opacity=0.3] table[x=Frame, y=estInst1]{\SAsequenceA};
        \addplot+[solid, A3, mark=none, fill opacity=0.3] table[x=Frame, y=estInst2]{\SAsequenceA};
        \addplot+[solid, A4, mark=none, fill opacity=0.3] table[x=Frame, y=estInst3]{\SAsequenceA};
        \addplot+[solid, C1, mark=none, fill opacity=0.3] table[x=Frame, y=estInst4]{\SAsequenceA};
        \addplot+[solid, C2, mark=none, fill opacity=0.3] table[x=Frame, y=estInst5]{\SAsequenceA};
    \end{axis}
    \end{tikzpicture}
    &
    \begin{tikzpicture}
    \begin{axis}[
        width=0.53\columnwidth,
        height=3.5cm,
        ymin=0,ymax=150,
        xmin=0,xmax=10000,
        label style={font=\footnotesize},
        tick label style={font=\footnotesize},
        ylabel near ticks,
        xlabel near ticks,
        xticklabels={,,},
        xtick scale label code/.code={},
    ]
        \addplot+[solid, intel, mark=none, style={thick}] table[x=Frame, y=actCum]{\SAsequenceA};
        \addplot+[solid, A3, mark=none, fill opacity=0.3] table[x=Frame, y=estCum2]{\SAsequenceA};
        \addplot+[solid, A4, mark=none, fill opacity=0.3] table[x=Frame, y=estCum3]{\SAsequenceA};
        \addplot+[solid, C1, mark=none, fill opacity=0.3] table[x=Frame, y=estCum4]{\SAsequenceA};
        \addplot+[solid, C2, mark=none, fill opacity=0.3] table[x=Frame, y=estCum5]{\SAsequenceA};
    \end{axis}
    \end{tikzpicture}
    \\
    \begin{tikzpicture}
    \begin{axis}[
        width=0.53\columnwidth,
        height=3.5cm,
        ylabel={{Airport-2}},
        ymin=0,ymax=20,
        xmin=0,xmax=10000,
        label style={font=\footnotesize},
        tick label style={font=\footnotesize},
        ylabel near ticks,
        xlabel near ticks,
        xticklabels={,,},
        xtick scale label code/.code={},
    ]
        \addplot+[solid, intel, mark=none, style={thick}] table[x=Frame, y=actInst]{\SAsequenceB};
        \addplot+[solid, A1, mark=none, fill opacity=0.3] table[x=Frame, y=estInst0]{\SAsequenceB};
        \addplot+[solid, A2, mark=none, fill opacity=0.3] table[x=Frame, y=estInst1]{\SAsequenceB};
        \addplot+[solid, A3, mark=none, fill opacity=0.3] table[x=Frame, y=estInst2]{\SAsequenceB};
        \addplot+[solid, A4, mark=none, fill opacity=0.3] table[x=Frame, y=estInst3]{\SAsequenceB};
        \addplot+[solid, C1, mark=none, fill opacity=0.3] table[x=Frame, y=estInst4]{\SAsequenceB};
        \addplot+[solid, C2, mark=none, fill opacity=0.3] table[x=Frame, y=estInst5]{\SAsequenceB};
    \end{axis}
    \end{tikzpicture}
    &
    \begin{tikzpicture}
    \begin{axis}[
        width=0.53\columnwidth,
        height=3.5cm,
        ymin=0,ymax=150,
        xmin=0,xmax=10000,
        label style={font=\footnotesize},
        tick label style={font=\footnotesize},
        ylabel near ticks,
        xlabel near ticks,
        xticklabels={,,},
        xtick scale label code/.code={},
    ]
        \addplot+[solid, intel, mark=none, style={thick}] table[x=Frame, y=actCum]{\SAsequenceB};
        \addplot+[solid, A3, mark=none, fill opacity=0.3] table[x=Frame, y=estCum2]{\SAsequenceB};
        \addplot+[solid, A4, mark=none, fill opacity=0.3] table[x=Frame, y=estCum3]{\SAsequenceB};
        \addplot+[solid, C1, mark=none, fill opacity=0.3] table[x=Frame, y=estCum4]{\SAsequenceB};
        \addplot+[solid, C2, mark=none, fill opacity=0.3] table[x=Frame, y=estCum5]{\SAsequenceB};
    \end{axis}
    \end{tikzpicture}
    \\
    \begin{tikzpicture}
    \begin{axis}[
        width=0.53\columnwidth,
        height=3.5cm,
        ylabel={{Mall-3}},
        ymin=0,ymax=20,
        xmin=0,xmax=10000,
        label style={font=\footnotesize},
        tick label style={font=\footnotesize},
        ylabel near ticks,
        xlabel near ticks,
        xticklabels={,,},
        xtick scale label code/.code={},
    ]
        \addplot+[solid, intel, mark=none, style={thick}] table[x=Frame, y=actInst]{\SAsequenceC};
        \addplot+[solid, A1, mark=none, fill opacity=0.3] table[x=Frame, y=estInst0]{\SAsequenceC};
        \addplot+[solid, A2, mark=none, fill opacity=0.3] table[x=Frame, y=estInst1]{\SAsequenceC};
        \addplot+[solid, A3, mark=none, fill opacity=0.3] table[x=Frame, y=estInst2]{\SAsequenceC};
        \addplot+[solid, A4, mark=none, fill opacity=0.3] table[x=Frame, y=estInst3]{\SAsequenceC};
        \addplot+[solid, C1, mark=none, fill opacity=0.3] table[x=Frame, y=estInst4]{\SAsequenceC};
        \addplot+[solid, C2, mark=none, fill opacity=0.3] table[x=Frame, y=estInst5]{\SAsequenceC};
    \end{axis}
    \end{tikzpicture}
    &
    \begin{tikzpicture}
    \begin{axis}[
        width=0.53\columnwidth,
        height=3.5cm,
        ymin=0,ymax=150,
        xmin=0,xmax=10000,
        label style={font=\footnotesize},
        tick label style={font=\footnotesize},
        ylabel near ticks,
        xlabel near ticks,
        xticklabels={,,},
        xtick scale label code/.code={},
    ]
        \addplot+[solid, intel, mark=none, style={thick}] table[x=Frame, y=actCum]{\SAsequenceC};
        \addplot+[solid, A3, mark=none, fill opacity=0.3] table[x=Frame, y=estCum2]{\SAsequenceC};
        \addplot+[solid, A4, mark=none, fill opacity=0.3] table[x=Frame, y=estCum3]{\SAsequenceC};
        \addplot+[solid, C1, mark=none, fill opacity=0.3] table[x=Frame, y=estCum4]{\SAsequenceC};
        \addplot+[solid, C2, mark=none, fill opacity=0.3] table[x=Frame, y=estCum5]{\SAsequenceC};
    \end{axis}
    \end{tikzpicture}
    \\
    \begin{tikzpicture}
    \begin{axis}[
        width=0.53\columnwidth,
        height=3.5cm,
        ylabel={{Mall-4}},
        ymin=0,ymax=20,
        xmin=0,xmax=10000,
        label style={font=\footnotesize},
        tick label style={font=\footnotesize},
        ylabel near ticks,
        xlabel near ticks,
        xticklabels={,,},
        xtick scale label code/.code={},
    ]
        \addplot+[solid, intel, mark=none, style={thick}] table[x=Frame, y=actInst]{\SAsequenceD};
        \addplot+[solid, A1, mark=none, fill opacity=0.3] table[x=Frame, y=estInst0]{\SAsequenceD};
        \addplot+[solid, A2, mark=none, fill opacity=0.3] table[x=Frame, y=estInst1]{\SAsequenceD};
        \addplot+[solid, A3, mark=none, fill opacity=0.3] table[x=Frame, y=estInst2]{\SAsequenceD};
        \addplot+[solid, A4, mark=none, fill opacity=0.3] table[x=Frame, y=estInst3]{\SAsequenceD};
        \addplot+[solid, C1, mark=none, fill opacity=0.3] table[x=Frame, y=estInst4]{\SAsequenceD};
        \addplot+[solid, C2, mark=none, fill opacity=0.3] table[x=Frame, y=estInst5]{\SAsequenceD};
    \end{axis}
    \end{tikzpicture}
    &
    \begin{tikzpicture}
    \begin{axis}[
        width=0.53\columnwidth,
        height=3.5cm,
        ymin=0,ymax=150,
        xmin=0,xmax=10000,
        label style={font=\footnotesize},
        tick label style={font=\footnotesize},
        ylabel near ticks,
        xlabel near ticks,
        xticklabels={,,},
        xtick scale label code/.code={},
    ]
        \addplot+[solid, intel, mark=none, style={thick}] table[x=Frame, y=actCum]{\SAsequenceD};
        \addplot+[solid, A3, mark=none, fill opacity=0.3] table[x=Frame, y=estCum2]{\SAsequenceD};
        \addplot+[solid, A4, mark=none, fill opacity=0.3] table[x=Frame, y=estCum3]{\SAsequenceD};
        \addplot+[solid, C1, mark=none, fill opacity=0.3] table[x=Frame, y=estCum4]{\SAsequenceD};
        \addplot+[solid, C2, mark=none, fill opacity=0.3] table[x=Frame, y=estCum5]{\SAsequenceD};
    \end{axis}
    \end{tikzpicture}
    \\
    \begin{tikzpicture}
    \begin{axis}[
        width=0.53\columnwidth,
        height=3.5cm,
        ymin=0,ymax=20,
        xmin=0,xmax=10000,
        ylabel={{Pedestrian-2}},
        label style={font=\footnotesize},
        tick label style={font=\footnotesize},
        ylabel near ticks,
        xlabel near ticks,
        xticklabels={,,},
        xtick scale label code/.code={},
    ]
        \addplot+[solid, intel, mark=none, style={thick}] table[x=Frame, y=actInst]{\SAsequenceE};
        \addplot+[solid, A1, mark=none, fill opacity=0.3] table[x=Frame, y=estInst0]{\SAsequenceE};
        \addplot+[solid, A2, mark=none, fill opacity=0.3] table[x=Frame, y=estInst1]{\SAsequenceE};
        \addplot+[solid, A3, mark=none, fill opacity=0.3] table[x=Frame, y=estInst2]{\SAsequenceE};
        \addplot+[solid, A4, mark=none, fill opacity=0.3] table[x=Frame, y=estInst3]{\SAsequenceE};
        \addplot+[solid, C1, mark=none, fill opacity=0.3] table[x=Frame, y=estInst4]{\SAsequenceE};
        \addplot+[solid, C2, mark=none, fill opacity=0.3] table[x=Frame, y=estInst5]{\SAsequenceE};
    \end{axis}
    \end{tikzpicture}
    &
    \begin{tikzpicture}
    \begin{axis}[
        width=0.53\columnwidth,
        height=3.5cm,
        ymin=0,ymax=150,
        xmin=0,xmax=10000,
        label style={font=\footnotesize},
        tick label style={font=\footnotesize},
        ylabel near ticks,
        xlabel near ticks,
        xticklabels={,,},
        xtick scale label code/.code={},
    ]
        \addplot+[solid, intel, mark=none, style={thick}] table[x=Frame, y=actCum]{\SAsequenceE};
        \addplot+[solid, A3, mark=none, fill opacity=0.3] table[x=Frame, y=estCum2]{\SAsequenceE};
        \addplot+[solid, A4, mark=none, fill opacity=0.3] table[x=Frame, y=estCum3]{\SAsequenceE};
        \addplot+[solid, C1, mark=none, fill opacity=0.3] table[x=Frame, y=estCum4]{\SAsequenceE};
        \addplot+[solid, C2, mark=none, fill opacity=0.3] table[x=Frame, y=estCum5]{\SAsequenceE};
    \end{axis}
    \end{tikzpicture}
    \\
    \begin{tikzpicture}
    \begin{axis}[
        width=0.53\columnwidth,
        height=3.5cm,
        ymin=0,ymax=20,
        xmin=0,xmax=10000,
        ylabel={{Pedestrian-3}},
        label style={font=\footnotesize},
        tick label style={font=\footnotesize},
        ylabel near ticks,
        xlabel near ticks,
        xlabel={Frames},
        tick scale binop=\times
    ]
        \addplot+[solid, intel, mark=none, style={thick}] table[x=Frame, y=actInst]{\SAsequenceF};
        \addplot+[solid, A1, mark=none, fill opacity=0.3] table[x=Frame, y=estInst0]{\SAsequenceF};
        \addplot+[solid, A2, mark=none, fill opacity=0.3] table[x=Frame, y=estInst1]{\SAsequenceF};
        \addplot+[solid, A3, mark=none, fill opacity=0.3] table[x=Frame, y=estInst2]{\SAsequenceF};
        \addplot+[solid, A4, mark=none, fill opacity=0.3] table[x=Frame, y=estInst3]{\SAsequenceF};
        \addplot+[solid, C1, mark=none, fill opacity=0.3] table[x=Frame, y=estInst4]{\SAsequenceF};
        \addplot+[solid, C2, mark=none, fill opacity=0.3] table[x=Frame, y=estInst5]{\SAsequenceF};
    \end{axis}
    \end{tikzpicture}
    &
    \begin{tikzpicture}
    \begin{axis}[
        width=0.53\columnwidth,
        height=3.5cm,
        ymin=0,ymax=150,
        xmin=0,xmax=10000,
        label style={font=\footnotesize},
        tick label style={font=\footnotesize},
        ylabel near ticks,
        xlabel near ticks,
        xlabel={Frames},
        tick scale binop=\times,
    ]
        \addplot+[solid, intel, mark=none, style={thick}] table[x=Frame, y=actCum]{\SAsequenceF};
        \addplot+[solid, A3, mark=none, fill opacity=0.3] table[x=Frame, y=estCum2]{\SAsequenceF};
        \addplot+[solid, A4, mark=none, fill opacity=0.3] table[x=Frame, y=estCum3]{\SAsequenceF};
        \addplot+[solid, C1, mark=none, fill opacity=0.3] table[x=Frame, y=estCum4]{\SAsequenceF};
        \addplot+[solid, C2, mark=none, fill opacity=0.3] table[x=Frame, y=estCum5]{\SAsequenceF};
    \end{axis}
    \end{tikzpicture}
    \end{tabular}
    \caption{Per-frame count of instantaneous and cumulative people with OTS performance in selected videos for
    Algorithm~1 (A1:~{\protect\raisebox{2pt}{\protect\tikz \protect\draw[A1,line width=1.5] (0,0) -- (0.3,0);}}),
    Algorithm~2 (A2:~{\protect\raisebox{2pt}{\protect\tikz \protect\draw[A2,line width=1.5] (0,0) -- (0.3,0);}}),
    Algorithm~3 (A3:~{\protect\raisebox{2pt}{\protect\tikz \protect\draw[A3,line width=1.5] (0,0) -- (0.3,0);}}),
    Algorithm~4 (A4:~{\protect\raisebox{2pt}{\protect\tikz \protect\draw[A4,line width=1.5] (0,0) -- (0.3,0);}}); and
    Commercial off-the-shelf~1 (C1:~{\protect\raisebox{2pt}{\protect\tikz \protect\draw[C1,line width=1.5] (0,0) -- (0.3,0);}}), and
    Commercial off-the-shelf~2 (C2:~{\protect\raisebox{2pt}{\protect\tikz \protect\draw[C2,line width=1.5] (0,0) -- (0.3,0);}}).
    The annotation count is shown in ({\protect\raisebox{2pt}{\protect\tikz \protect\draw[intel,line width=1.5] (0,0) -- (0.3,0);}}).
    All algorithms have been executed with the same video inputs (at 30~frames per second).
    A1-4 are run in System 1 with GPU. C1-2 are run in CPU.
    For easing the visualization, the cumulative count of A1-2 are not shown (they are several orders of magnitude larger than the rest of the algorithms); and, only one sample per second (instead of 30 samples per second) is shown. Note that the y axes are limited to 20 (instantaneous) and 150 (cumulative), and that x axes are limited to 10000 frames.}
\label{fig:perframe}
\end{figure}